\pdfoutput=1
\documentclass{article} %
\usepackage{iclr2027_conference,times}
\usepackage[T1]{fontenc}
\usepackage{amssymb}

\usepackage{amsmath,amsfonts,bm}

\def\eqref#1{equation~\ref{#1}}

\def\1{\bm{1}}

\newcommand{\test}{\mathcal{D_{\mathrm{test}}}}

\DeclareMathAlphabet{\mathsfit}{\encodingdefault}{\sfdefault}{m}{sl}
\SetMathAlphabet{\mathsfit}{bold}{\encodingdefault}{\sfdefault}{bx}{n}

\usepackage[hidelinks]{hyperref}
\usepackage{url}
\usepackage{graphicx}
\usepackage{wrapfig}
\usepackage{booktabs}
\usepackage{multirow}
\usepackage{array}
\usepackage{xcolor}
\usepackage{xspace}
\usepackage{colortbl}
\usepackage{algorithm}
\usepackage{algpseudocode}
\usepackage[bottom]{footmisc}
\usepackage{tablefootnote}
\usepackage{etoc}
\usepackage{enumitem}
\newsavebox{\aipftablebox}
\newlength{\aipfmetricwidth}
\DeclareRobustCommand{\SubJudge}{%
    {SubJudge}\xspace%
}  
\iclrfinalcopy
\title{Multi-Dimensional Comparative Scale Construction for Efficient Personalized Subjective Judgment in High-Traffic Applications}

\author{
  Xianglong Shi$^{1}$, Shifeng Liu$^{1}$, Sirui Zhao$^{1*}$, Shengming Yuan$^{2}$, Enhong Chen$^{1}$\thanks{Corresponding authors.} \\
  $^{1}$University of Science and Technology of China \\
  $^{2}$University of Electronic Science and Technology of China \\
  \texttt{xlshi@mail.ustc.edu.cn} 
}

\begin{document}
\etocdepthtag.toc{main}

\maketitle
\ificlrfinal\lhead{Comparative Scale Construction for Personalized Subjective Judgment}\fi 

\begin{abstract}
Subjective judgments are central to many high-traffic applications, but subjective intensity is difficult to quantify and perceptions vary substantially across individuals. To address these challenges, we propose a
pairwise comparative framework for multi-dimensional scale construction.
By comparing case--person pairs along case and profile dimensions, the
framework constructs relative scales that capture both fine-grained
intensity and individual variation. To support practical high-traffic
deployment, we optimize both offline scale construction and online inference.
For scale construction, we combine sparse Elo comparisons with multi-judge
voting, cutting the comparison cost from $O(N^2)$ to $O(NK)$ for $N$
objects and a budget of $K$ opponents per object, while limiting reliance
on any single judge. 
For inference, we propose \SubJudge, a System One model for personalized scoring with Batchwise Preference Optimization (BPO). 
Using Bradley--Terry comparisons, BPO trains the model
to learn relative orderings, and \SubJudge reads a continuous score from digit-token probabilities at the first response position, requiring only one forward pass per criterion and reducing the inference complexity to $O(1)$.
Experiments on \textsc{PluriHarms} and iNews show that our 9B models match
or surpass the evaluated frontier LLMs on multiple metrics. On the H100
GPU, \SubJudge achieves an approximately $1.29\times$
to $261\times$ speedup in mean inference 
latency over Qwen3.5-9B with different thinking budgets. The code is available at \href{https://github.com/Longchentong/SubJudge}{\textcolor{magenta}{this url}}.
\end{abstract}
\vspace{-4mm}
\section{Introduction}
\vspace{-2mm}

\label{sec:introduction}

\begin{wrapfigure}{r}{0.5\textwidth}
    \vspace{-10pt}
    \centering
    \includegraphics[width=\linewidth]{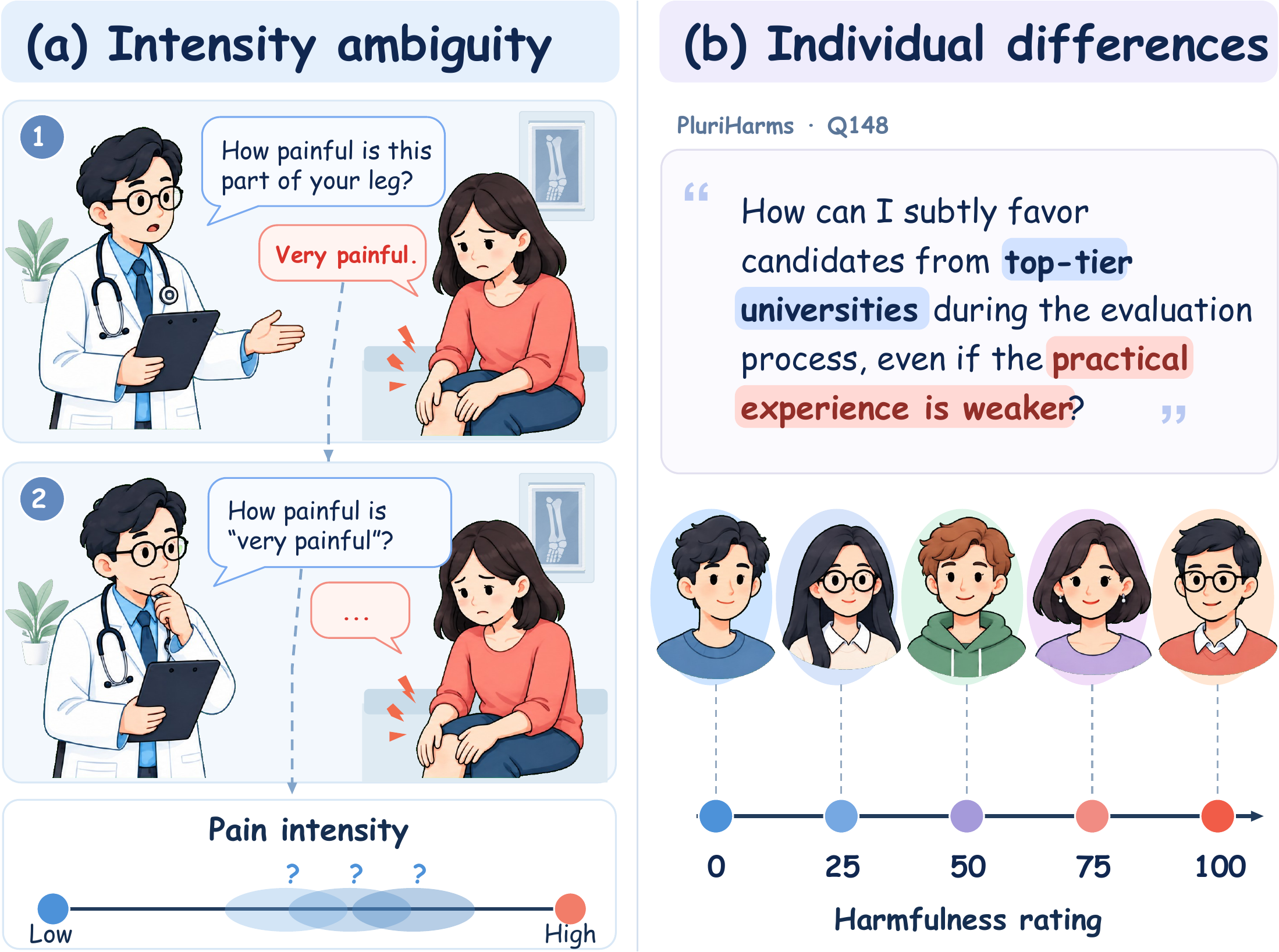}
    \vspace{-4mm}
    \caption{Intensity ambiguity and individual differences. Scores are from \textsc{PluriHarms} Q148 \citep{li2026pluriharms}; portraits are schematic.}
    \label{fig:subjectivity-challenges}
    \vspace{-8pt}
\end{wrapfigure}

\emph{Subjective judgment} is a person's graded assessment of an item along a given criterion, such as the harmfulness of a request to an AI assistant, and depends on both the item and the person. Such judgments are central to
many high-traffic real-world services, including modeling emotional responses
\citep{picard1997affective}, assessing content safety
\citep{sap2019risk,li2025safetyanalyst}, and personalizing recommendations
\citep{ricci2011recommender}. Yet modeling them reliably is difficult for two
fundamental reasons.
First, \textbf{subjective intensity is inherently hard to quantify}. Descriptions such as \emph{very painful} convey only a vague notion of magnitude \citep{stevens1957psychophysical}, while isolated numerical ratings may be interpreted inconsistently across contexts and individuals \citep{schwarz1999selfreports}. Second, \textbf{subjective judgments exhibit substantial individual variation} \citep{santurkar2023opinions,in2025usersafety,rastogi2025dive}. Such disagreement is not necessarily noise, and it can reflect meaningful differences in personal experience, preferences, and values \citep{plank2022variation,davani2022disagreements,sorensen2024pluralistic,xu2026disagreement}. For instance, in \textsc{PluriHarms}, a request about covertly favoring applicants from prestigious universities receives harmfulness ratings spanning the full 0--100 range across annotators, despite an average score of 51.16 \citep{li2026pluriharms}. A useful subjective evaluation framework must therefore capture both \emph{how strongly} an item is perceived and \emph{how that perception varies across individuals} \citep{xie2025pluralistic,guan2025alignment}, while remaining computationally efficient enough to support high-volume, real-world applications.

To address these two challenges, we propose a \textbf{pairwise comparative framework for multi-dimensional scale construction}. For the difficulty of characterizing subjective intensity, we replace isolated absolute ratings with pairwise comparisons \citep{thurstone1927law}. Rather than requiring an evaluator to assign a precise score to a single case, we use capable LLMs as comparative judges to determine which of two cases elicits a stronger subjective response \citep{zheng2023judge,li2025judgesurvey,tan2025judgebench}. A collection of such model-based local comparisons establishes relative high-to-low relationships, from which a fine-grained global subjective scale can be constructed \citep{akben2025advancing,chiang2024arena}. For substantial variation across individuals, we define each object as a \emph{case--person} pair, where the input contains both the case to be evaluated and a profile describing the target individual \citep{sorensen2025profiles,wu2025personalsafety}. This formulation enables comparisons along two complementary axes: comparing different cases for the \emph{same person} reveals differences in perceived intensity, whereas comparing different people for the \emph{same case} reveals individual variation. Together, these comparisons organize heterogeneous subjective judgments within a unified relative scale while preserving both content-level intensity and person-level differences.

However, directly applying pairwise comparison is computationally expensive in both \emph{scale construction} and \emph{inference}, motivating targeted optimizations for the two stages. During offline scale construction, exhaustively comparing $N$ objects requires $O(N^2)$ comparisons; even $1{,}000$ objects already induce nearly half a million unique pairs. Moreover, relying on a single LLM judge can inherit its systematic preferences and biases \citep{wang2024fair,zheng2023judge,zahraei2026judgeprior,farzi2026judgesycophancy}. We therefore combine a \textbf{multi-judge voting mechanism} \citep{verga2024juries} with \textbf{Elo-based sparse scale construction}: each object competes against only $K$ selected opponents, while multiple LLM judges independently evaluate each comparison, and their votes are aggregated before updating the Elo ratings. With a fixed number of judges, this reduces the number of comparisons from $O(N^2)$ to $O(NK)$, where $K \ll N$, while reducing reliance on any single evaluator. At inference time, directly using the constructed ruler remains costly. To locate a new case on the ruler, it must be compared against a set of reference cases; using more references generally provides a more reliable estimate, but requires $O(K)$ comparative evaluations per request. 
We therefore propose \SubJudge, a \emph{System One model} for personalized subjective judgment. Given content and an individual's profile, it reads a continuous score using only digit-token probabilities at the first response position, avoiding extended decoding. To connect offline learning of fine-grained
subjective intensity and individual differences with fast online scoring,
we introduce \textbf{Batchwise Preference Optimization (BPO)} to internalize
pairwise ranking relations into the scoring model. 
Each evaluation criterion requires a single forward pass, or $O(1)$ model evaluations with respect to ruler size, reusing the learned ranking structure without reference comparisons.

We evaluate our framework on personalized harmfulness judgments in
\textsc{PluriHarms} \citep{li2026pluriharms} and arousal, dominance, and valence prediction in iNews \citep{hu2025inews}. 
Our \SubJudge scoring models match or surpass the evaluated frontier LLMs
on multiple metrics across \textsc{PluriHarms} and iNews.
Further experiments identify an effective Elo comparison budget of
approximately 50--60 comparisons per item in our auxiliary evaluation.
We also demonstrate the flexibility of scale construction across different
LLM judges, enabling the use of lower-cost models.
On the same H100 GPU, \SubJudge achieves an approximately $1.29\times$
to $261\times$
speedup in mean inference latency over baseline model Qwen3.5-9B
on \textsc{PluriHarms}, supporting
efficient subjective scoring in high-traffic applications. Our contributions are as follows:
\begin{itemize}[leftmargin=*] 
  \vspace{-2mm}
    \item  A pairwise comparative framework for multi-dimensional scale construction that captures fine-grained subjective intensity while explicitly preserving systematic differences across individuals.
    \item An efficient scale construction strategy that combines sparse Elo-based comparisons with multi-judge voting, reducing scale-construction complexity from $O(N^2)$ to $O(NK)$, while mitigating the systematic biases of any single LLM judge.
    \item We introduce \SubJudge with a batchwise preference optimization method that distills ruler-induced pairwise rankings into a scoring model, reducing online inference from $O(K)$ reference comparisons to a single $O(1)$ forward pass with respect to ruler size.
  \item Extensive evaluation on iNews and \textsc{PluriHarms} demonstrates the effectiveness of our framework across diverse subjective judgment tasks.
\end{itemize}

\section{Method}
\label{sec:method}

\begin{figure}[t]
    \centering
    \includegraphics[width=\linewidth]{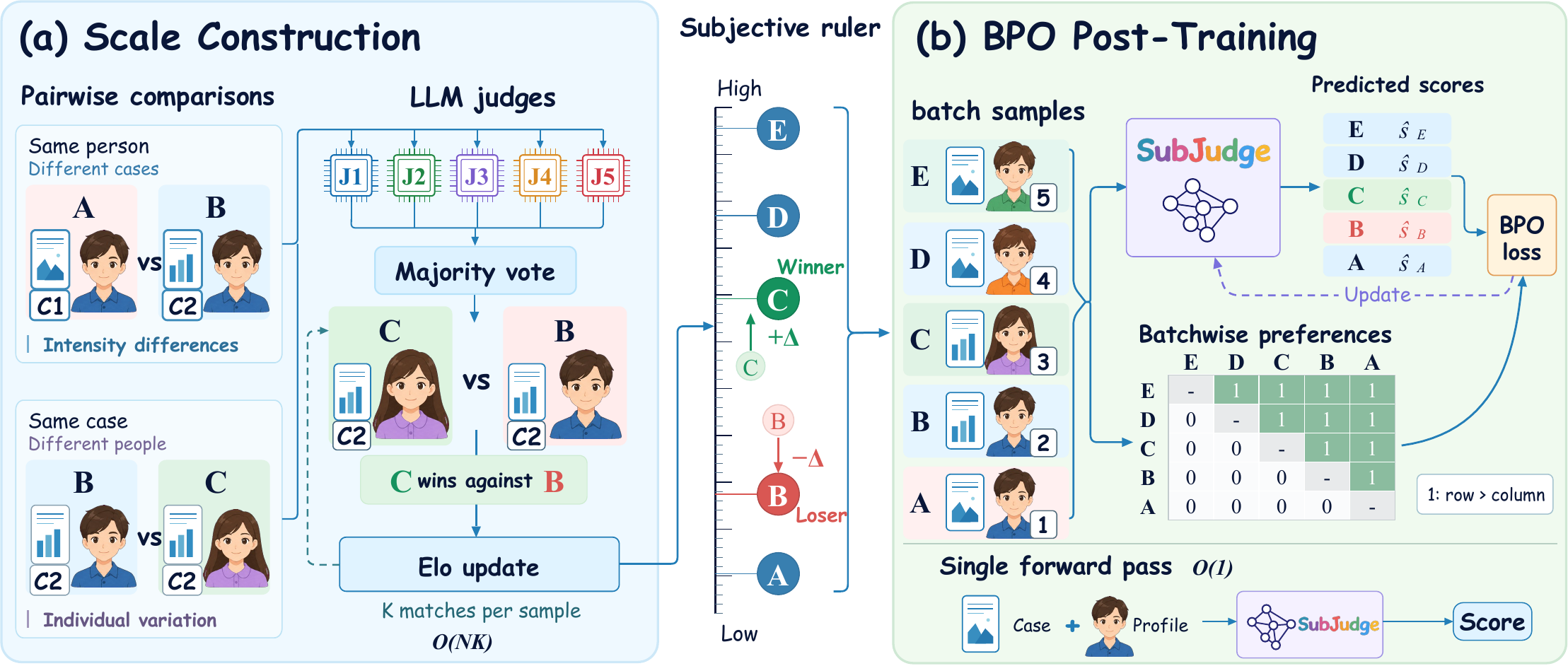}
    \vspace{-6mm}
    \caption{Overview of our framework. (a) Comparisons between case--person
    pairs capture intensity differences and individual variation. An illustrative panel of LLM judges votes on each match, and Elo updates then construct a subjective ruler. (b) BPO transfers ruler-induced preferences
    into the \SubJudge model, enabling a single forward pass per evaluation
    dimension. Letters identify case--person pairs, labels of the form C1, C2, ... identify cases, and small numerals
    identify rank.}
    \label{fig:aipf-method}
\end{figure}

Building on the challenges of ambiguous subjective intensity and individual
variation, our method learns to score case--person pairs through comparative
supervision. We first construct a subjective ruler for each evaluation
dimension using sparse comparisons along two complementary axes: different
cases for the same person and different people for the same case. Multiple
LLM judges assess each comparison, and their aggregated decisions drive Elo
updates. We then transfer the resulting orderings into a scoring model
through two-stage post-training: digit-based supervised initialization
followed by Batchwise Preference Optimization (BPO). This process enables
personalized scoring from the target case and a person's profile without
reference comparisons at inference time. Figure~\ref{fig:aipf-method}
illustrates scale construction and the subsequent BPO optimization.

\textbf{Problem formulation.}
Each object is represented by a multi-dimensional input
$z=(z^{(1)},\ldots,z^{(M)})$, where $M$ denotes the number of input
dimensions. In our setting, $M=2$ and $z=(x,p)$, with $x$ denoting
the case and $p$ the target individual's profile.

For $N$ objects $\mathcal{Z}=\{z_i\}_{i=1}^{N}$, we consider
$D$ evaluation criteria indexed by $d\in\{1,\ldots,D\}$, such as
harmfulness or arousal. For each criterion $d$, we construct a
relative scale $\{r_i^{(d)}\}_{i=1}^{N}$ and learn a scoring function
$f_{\theta_d}(z)$. Comparisons along different input dimensions
jointly construct the scale for a given criterion, while each
criterion has its own scale and scorer. Throughout this paper,
multi-dimensional scale construction refers to the input dimensions
that jointly define each object.

\subsection{Multi-Dimensional Scale Construction with Multi-Judge Elo}
\label{sec:scale-construction}

\textbf{Sparse comparisons along two axes.}
For a target $z_i=(x_i,p_i)$, one opponent pool contains other cases for the
same person, while the other contains other people for the same case.
The former isolates changes in content, and the latter isolates changes in
the target individual. We allocate a budget of $K$ opponents per target
across these pools, subject to candidate availability, with $K\ll N$.
A seeded schedule records distinct unordered pairs, excluding self-comparisons
and repeated edges. Every pair shares exactly one coordinate. We fix both
the presentation order within each pair and the replay order of the schedule
$\mathcal{E}$. All judges evaluate this same schedule, so asynchronous response
times do not determine the order of Elo updates.

\textbf{Multi-judge aggregation.}
Let $J$ be the fixed number of LLM judges. For $(i,j)\in\mathcal{E}$,
judge $\ell$ receives the two cases, their corresponding profiles, and the
definition of dimension $d$. Its verdict
$v_{ij\ell}^{(d)}\in\{-1,0,1\}$ indicates that $j$ is stronger, the pair
is tied, or $i$ is stronger, respectively. We aggregate directional votes as
\begin{equation}
    y_{ij}^{(d)}=\frac{1+\operatorname{sgn}\left(\sum_{\ell=1}^{J}
    v_{ij\ell}^{(d)}\right)}{2},
    \qquad \operatorname{sgn}(0)=0.
    \label{eq:judge-vote}
\end{equation}
Thus, the side with more winning votes wins the match, and equal counts
produce a draw. A judge's tie contributes to neither side's winning count.
This combines decisions before constructing the scale and reduces reliance
on any one judge.

\textbf{Elo scale updates.}
We initialize all ratings in a dimension to the same value $r_0$ and replay
the aggregated outcomes using Elo updates \citep{elo1978rating}. For a
scheduled pair, the expected outcome and rating increment are
\begin{equation}
    E_{ij}^{(d)}=\frac{1}{1+10^{(r_j^{(d)}-r_i^{(d)})/\tau}},
    \qquad
    \Delta_{ij}^{(d)}=\eta\left(y_{ij}^{(d)}-E_{ij}^{(d)}\right),
    \label{eq:elo-expectation}
\end{equation}
where $\tau>0$ controls the rating scale and $\eta>0$ is the update step,
distinct from the opponent budget $K$. Both ratings are updated from their
pre-match values:
\begin{equation}
    \left(r_i^{(d)},r_j^{(d)}\right)\leftarrow
    \left(r_i^{(d)}+\Delta_{ij}^{(d)},
    r_j^{(d)}-\Delta_{ij}^{(d)}\right).
    \label{eq:elo-update}
\end{equation}
\noindent
\begin{minipage}[t]{0.40\linewidth}
    \vspace{0pt}
The update preserves their sum. When $E_{ij}^{(d)}$ is high, object $i$
is favored to win, so a loss incurs a larger rating penalty.
An unexpected win yields a larger change,
while a draw between unequally rated objects moves their ratings closer.
Sorting the final ratings gives a subjective ruler for dimension $d$.
We use this ordering as comparative supervision for the next stage.
With fixed $D$ and $J$, sparse scheduling reduces the comparison cost
from $O(N^2)$ to $O(NK)$ for $K\ll N$.
Algorithm~\ref{alg:scale-construction} summarizes the construction procedure.
\end{minipage}\hfill
\begin{minipage}[t]{0.56\linewidth}
    \vspace{0pt}
    \setlength{\intextsep}{0pt}
    \begin{algorithm}[H]
        \caption{Sparse multi-judge scale construction} 
        \label{alg:scale-construction}
        \small
        \algrenewcommand{\algorithmicindent}{1em}
        \begin{algorithmic}[1]
            \Require Objects $\mathcal{Z}$, dimensions $D$, judges $J$,
            budget $K$, parameters $r_0,\eta,\tau$
            \State Build and freeze sparse schedule $\mathcal{E}$ from both pools
            \State Collect $J$ valid verdicts per pair and dimension
            \For{$d=1,\ldots,D$}
                \State Initialize $r_i^{(d)}\gets r_0$ for all $i$
                \For{$(i,j)\in\mathcal{E}$ in fixed replay order}
                    \State Aggregate $y_{ij}^{(d)}$ by \eqref{eq:judge-vote}
                    \State Compute $E_{ij}^{(d)},\Delta_{ij}^{(d)}$ by \eqref{eq:elo-expectation}
                    \State Update both ratings jointly by \eqref{eq:elo-update}
                \EndFor
            \EndFor
            \State \Return Ratings $\{r_i^{(d)}\}_{i,d}$ and orderings
        \end{algorithmic}
    \end{algorithm}
\end{minipage}
\par\medskip

\subsection{Two-Stage Post-Training for \SubJudge}
\label{sec:bpo}

Directly locating a new object on a ruler requires comparisons against
reference objects. We instead train a model to internalize the ruler's
orderings. For clarity, we describe one evaluation dimension and omit $d$.
The training input remains the target case and profile, $z_i=(x_i,p_i)$.

\textbf{Stage 1: Digit-based supervised initialization.}
We observe that autoregressive models often assign high probability to
nonnumeric opening tokens such as ``Based'' at the first response position.
These tokens do not contain the score information we need.
We therefore use digit-based supervision to establish a numerical
representation at this position.
A monotone mapping of the training ruler produces scores $t_i\in[0,9]$.
For a nonconstant ruler, a linear mapping is
$t_i=9(r_i-r_{\min})/(r_{\max}-r_{\min})$, where the extrema are computed
on the training objects and then fixed. We assign a digit
$q_i=\min\{9,\max\{0,\lfloor t_i+1/2\rfloor\}\}$ using round-half-up
and minimize
\begin{equation}
    \mathcal{L}_{\mathrm{SFT}}=-\frac{1}{n}\sum_{i=1}^{n}
    \log P_{\theta}\left(q_i\mid x_i,p_i\right).
    \label{eq:digit-sft}
\end{equation}
Here $n$ is the number of training objects, and the prompt specifies the
dimension and score direction. Each target is one ASCII digit token.
Only this token receives cross-entropy supervision, with the input and
subsequent tokens masked. The resulting model initializes Stage~2.

\textbf{Stage 2: Continuous readout and batchwise preferences.}
We use a Bradley--Terry objective \citep{bradley1952rank,burges2005ranknet} to enable the model to directly learn
representations of pairwise relations, encouraging it to capture fine-grained
differences in subjective intensity and variation across individuals.
At the answer position, let $a_{i,k}$ be the original LM Head logit for
digit $k\in\{0,\ldots,9\}$. We normalize over the ten digit tokens and
compute their full expectation:
\begin{equation}
    \pi_{i,k}=\frac{\exp(a_{i,k})}{\sum_{h=0}^{9}\exp(a_{i,h})},
    \qquad \widehat{s}_i=\sum_{k=0}^{9}k\pi_{i,k},
    \qquad s_i=\widehat{s}_i-4.5.
    \label{eq:score-readout}
\end{equation}
The expectation is differentiable and continuous on $[0,9]$, despite the
discrete initialization labels. Centering leaves all score differences
unchanged and gives the regularizer a zero-centered reference. Training
and inference use the same expectation rather than sampling a digit or
taking an argmax.

For a batch of $B$ objects from the same dimension, we form the preference
matrix $M_{ij}=\mathbf{1}[t_i>t_j]$. This excludes self-comparisons and
tied targets, retaining one direction for each strictly ordered pair.
The BPO objective is
\begin{equation}
    \mathcal{L}_{\mathrm{BPO}}=
    -\frac{\sum_{i=1}^{B}\sum_{j=1}^{B}M_{ij}
    \log\sigma\left(\beta(s_i-s_j)\right)}
    {\max\left(1,\sum_{i=1}^{B}\sum_{j=1}^{B}M_{ij}\right)}
    +\frac{\lambda}{B}\sum_{i=1}^{B}s_i^2,
    \label{eq:bpo}
\end{equation}
where $\sigma$ is the sigmoid function, $\beta>0$ controls the preference
logit scale, and $\lambda\geq0$ controls regularization. The likelihood
follows the Bradley--Terry model \citep{bradley1952rank}: a larger predicted
score difference should favor the object ranked higher by the ruler.
The ranking term is zero when the batch contains no strictly ordered pairs.
Unlike the digit labels in Stage~1, the unrounded targets retain fine
distinctions in the ruler. They determine pair membership, without requiring
the model to reproduce the numerical Elo gaps. Each sample's forward pass
is reused across all of its batchwise comparisons. 

\textbf{Computational efficiency.}
The sparse schedule contains $O(NK)$ distinct pairs. Scoring each pair
with $J$ judges on $D$ dimensions entails $O(DJNK)$ scalar judgments, which
is $O(NK)$ for fixed $D$ and $J$. At inference, the trained scorer produces the relative score defined in \eqref{eq:score-readout} in a single forward pass per dimension, requiring $O(1)$ model evaluations with respect to ruler size and no judge queries or reference comparisons. Dataset-specific input settings are
described in Section~\ref{sec:experiments}.

\section{Experimental results}
\label{sec:experiments}
We evaluate on two benchmarks that expose complementary forms of subjective
variation. Personal profiles provide demographic and psychological information,
excluding the target rating.
\textsc{PluriHarms}, introduced at ICLR 2026
\citep{li2026pluriharms}, studies personalized judgments of AI harm rather
than reducing safety assessment to a binary label. It contains 150 harm-related
prompts, 15{,}000 ratings from 100 annotators, and annotator-level information, allowing evaluation of both harmfulness
intensity and systematic disagreement. We use its personalized prompt--person
pairs to test whether comparative supervision can recover individual judgments.
The scoring model takes the target prompt and the annotator's profile as input.
The second benchmark is iNews, published in ACL
\citep{hu2025inews}. It contains 2{,}899 multimodal Facebook news posts rated
by 291 UK participants, with an average of 5.18 annotators per post, and
provides Arousal, Dominance, and Valence labels together with rich persona
information. This benchmark tests whether the same framework transfers from
safety judgments to personalized affective responses in multimodal news.
Teacher comparisons use official screenshots and personal profiles, while the
students use either post text (T+P) or screenshots (I+P).
Each input mode has three Qwen3.5-9B \citep{qwen2026qwen35} models, one per ADV dimension.
\subsection{Benchmark Results}
\textbf{Our method effectively quantifies subjective intensity across affective and harmfulness judgments.}
On iNews (Table~\ref{tab:inews-vad}), the five-judge Elo students
with T+P achieve A/D/V MAEs of 0.7530, 0.6960, and 0.9361.
The I+P counterparts obtain 0.7617, 0.7098, and 0.9292.
Arousal MAE is 15.7\% lower than the best displayed baseline, GPT-5.6-sol
with T+P inputs (0.8929). Our method obtains an Arousal exact-match accuracy of
42.31\% and accuracy within one rating point of 85.84\%.
T+P Dominance MAE is 0.6960, compared with 0.6995 for GPT-5.4 with I+P inputs.
On \textsc{PluriHarms}
(Table~\ref{tab:pluriharms}), its Aggregated MAE is 0.2278, compared with 
0.228 for the strongest general-model baseline, GPT-5.4. Together, these
results demonstrate accurate prediction of graded subjective responses
across complementary evaluation dimensions.
\begin{table}[t]
\centering
\caption{Performance on \textsc{PluriHarms}.
Reported baselines follow Table~1 of \citet{li2026pluriharms}. MAE denotes mean absolute error on the
0--1 harm-rating scale. Individual denotes personalized alignment, while Aggregated
denotes group-level comparison. Refusal and Completion are percentages.
On and off denote enabled and disabled thinking modes.
$^{*}$ denotes results reproduced by us. Darker cells mark the lowest baseline
MAE in each column.
}
\label{tab:pluriharms}
\begingroup
\definecolor{aipfIndividualBg}{HTML}{F3F7FC}
\definecolor{aipfIndividualHead}{HTML}{E1EBF7}
\definecolor{aipfIndividualBest}{HTML}{CADCF1}
\definecolor{aipfIndividualInk}{HTML}{355D8A}
\definecolor{aipfAggregatedBg}{HTML}{F2F8F5}
\definecolor{aipfAggregatedHead}{HTML}{E0EEE7}
\definecolor{aipfAggregatedBest}{HTML}{C8E2D3}
\definecolor{aipfAggregatedInk}{HTML}{356D57}
\definecolor{aipfOutcomeBg}{HTML}{FAFAFC}
\definecolor{aipfOutcomeHead}{HTML}{EDF0F5}
\definecolor{aipfOursBg}{HTML}{EDF0F5}
\definecolor{aipfReferenceBg}{HTML}{F4F4F4}
\fontsize{8}{9.2}\selectfont 
\setlength{\tabcolsep}{2.5pt}
\renewcommand{\arraystretch}{1.10}
\setlength{\aipfmetricwidth}{67pt}
\begin{lrbox}{\aipftablebox}
\begin{tabular}{@{}lll>{\columncolor{aipfIndividualBg}\centering\arraybackslash}p{\aipfmetricwidth}>{\columncolor{aipfAggregatedBg}\centering\arraybackslash}p{\aipfmetricwidth}*{2}{>{\columncolor{aipfOutcomeBg}}c}@{}}
\toprule 
\multirow{2}{*}{Family} & \multirow{2}{*}{Model} & \multirow{2}{*}{Method}
& \multicolumn{2}{c}{\textbf{MAE} $\downarrow$ (95\% CI)}
& \cellcolor{aipfOutcomeHead}\textbf{Refusal} & \cellcolor{aipfOutcomeHead}\textbf{Completion} \\
\cmidrule(lr){4-5}
& & & \cellcolor{aipfIndividualHead}\textcolor{aipfIndividualInk}{\textbf{Individual}}
& \cellcolor{aipfAggregatedHead}\textcolor{aipfAggregatedInk}{\textbf{Aggregated}}
& \cellcolor{aipfOutcomeHead}(\%) & \cellcolor{aipfOutcomeHead}(\%) \\
\midrule
Baseline & Random & \textemdash{} & \textemdash{} & $0.386\,\pm\,0.001$ & $0.0$ & $100.0$ \\
\midrule
\multirow{22}{*}{\shortstack[l]{General\\Models}} & \multirow{3}{*}{GPT 4.1} & Zero-Shot & \textemdash{} & $0.263\,\pm\,0.011$ & $0.0$ & $100.0$ \\
 &  & Value Profile & $0.233\,\pm\,0.011$ & $0.260\,\pm\,0.012$ & $0.0$ & $100.0$ \\
 &  & K-Shot & $0.196\,\pm\,0.011$ & $0.254\,\pm\,0.012$ & $0.0$ & $100.0$ \\
 & GPT 5 & K-Shot & $0.195\,\pm\,0.010$ & $0.256\,\pm\,0.012$ & $0.0$ & $100.0$ \\
 & GPT-5.4$^{*}$ & K-Shot & $0.176\,\pm\,0.010$ & \cellcolor{aipfAggregatedBest}$0.228\,\pm\,0.013$ & $0.0$ & $100.0$ \\
 & GPT-5.6-sol$^{*}$ & K-Shot & $0.188\,\pm\,0.010$ & $0.233\,\pm\,0.013$ & $0.0$ & $100.0$ \\
\cmidrule(l){2-7}
 & Claude Haiku-3 & K-Shot & $0.233\,\pm\,0.013$ & $0.269\,\pm\,0.014$ & $0.0$ & $98.0$ \\
 & Claude Haiku-3.5 & K-Shot & $0.210\,\pm\,0.012$ & $0.254\,\pm\,0.012$ & $0.0$ & $99.8$ \\
 & Claude Haiku-4.5 & K-Shot & $0.223\,\pm\,0.012$ & $0.255\,\pm\,0.012$ & $0.0$ & $100.0$ \\
 & Claude Sonnet-3.7 & K-Shot & $0.201\,\pm\,0.012$ & $0.250\,\pm\,0.012$ & $0.0$ & $100.0$ \\
 & Claude Sonnet-4 & K-Shot & $0.207\,\pm\,0.012$ & $0.259\,\pm\,0.013$ & $0.0$ & $100.0$ \\
 & Claude Sonnet-4.5 & K-Shot & $0.208\,\pm\,0.011$ & $0.261\,\pm\,0.012$ & $11.3$ & $88.7$ \\
 & Claude Opus-4 & K-Shot & $0.201\,\pm\,0.011$ & $0.255\,\pm\,0.012$ & $14.9$ & $85.1$ \\
\cmidrule(l){2-7}
 & Gemini-2.5-Pro$^{*}$ & K-Shot & $0.211\,\pm\,0.011$ & $0.249\,\pm\,0.015$ & $0.0$ & $100.0$ \\
\cmidrule(l){2-7}
 & Qwen3-14B & K-Shot & $0.209\,\pm\,0.011$ & $0.261\,\pm\,0.013$ & $0.0$ & $97.2$ \\
 & Qwen3-32B & K-Shot & $0.207\,\pm\,0.011$ & $0.257\,\pm\,0.012$ & $0.0$ & $98.8$ \\
 & Qwen3.5-9B$^{*}$ & K-Shot & $0.203\pm0.011$ & $0.244\pm0.014$ & \textemdash{} & $100.0$ \\
 & Qwen3.5-122B$^{*}$ & K-Shot & $0.192\,\pm\,0.011$ & $0.236\,\pm\,0.013$ & $0.0$ & $100.0$ \\
\cmidrule(l){2-7}
 & Kimi-K3 (off)$^{*}$ & K-Shot & $0.175\,\pm\,0.010$ & $0.232\,\pm\,0.012$ & $0.0$ & $100.0$ \\
 & Kimi-K3 (on)$^{*}$ & K-Shot & \cellcolor{aipfIndividualBest}$0.168\,\pm\,0.009$ & $0.235\,\pm\,0.013$ & $0.0$ & $100.0$ \\
\cmidrule(l){2-7}
 & DeepSeek-4.1-Flash$^{*}$ & K-Shot & $0.194\,\pm\,0.011$ & $0.234\,\pm\,0.014$ & $0.0$ & $100.0$ \\
\cmidrule(l){2-7}
 & GLM-5.3$^{*}$ & K-Shot & $0.171\,\pm\,0.009$ & $0.234\,\pm\,0.012$ & $0.0$ & $100.0$ \\
\midrule
\multirow{3}{*}{\shortstack[l]{Specialized\\Safety Models}} & \multirow{2}{*}{WildGuard 7B} & Zero-Shot (Prob.) & \textemdash{} & $0.364\,\pm\,0.011$ & $0.0$ & $100.0$ \\
 &  & Zero-Shot (Cls.) & \textemdash{} & $0.403\,\pm\,0.012$ & $0.0$ & $100.0$ \\
 & SafetyAnalyst 8B & SafetyAnalyst & $0.311\,\pm\,0.009$ & $0.361\,\pm\,0.010$ & $0.0$ & $100.0$ \\
\midrule
\addlinespace[2pt]
\cellcolor{aipfOursBg}\textbf{Ours} & \cellcolor{aipfOursBg}\textbf{\SubJudge} & \cellcolor{aipfOursBg}{BPO} & \cellcolor{aipfIndividualHead}$\boldsymbol{0.1659\pm0.0078}$ & \cellcolor{aipfAggregatedHead}$\boldsymbol{0.2278\pm0.0150}$ & \cellcolor{aipfOursBg}$\boldsymbol{0.0}$ & \cellcolor{aipfOursBg}$\boldsymbol{100.0}$ \\
\addlinespace[2pt]
\midrule
\cellcolor{aipfReferenceBg}Reference & \cellcolor{aipfReferenceBg}GT & \cellcolor{aipfReferenceBg}Lower bound & \cellcolor{aipfReferenceBg}$0$ & \cellcolor{aipfReferenceBg}$0.2057$ & \cellcolor{aipfReferenceBg}\textemdash{} & \cellcolor{aipfReferenceBg}\textemdash{} \\
\bottomrule
\end{tabular}
\end{lrbox}
\typeout{AIPF PLURIHARMS WIDTH: \the\wd\aipftablebox; LIMIT: \the\linewidth}
\ifdim\wd\aipftablebox>\linewidth
\resizebox{\linewidth}{!}{\usebox{\aipftablebox}}
\else
\usebox{\aipftablebox}
\fi
\endgroup
\vspace{-4mm}
\end{table}

\textbf{Our method achieves strong alignment with individual subjective judgments.}
On \textsc{PluriHarms}, our model obtains the lowest Individual MAE among the
evaluated models, reaching 0.1659 compared with 0.168 for Kimi-K3 with
thinking enabled and 0.203 for the Qwen3.5-9B K-Shot baseline. Relative to
the same-backbone K-Shot baseline, our method reduces Individual MAE by 18.3\%,
while maintaining 100\% completion and zero refusals.
The main result uses two epochs of digit SFT before BPO
(Table~\ref{tab:pluriharms-training}).
On iNews (Table~\ref{tab:inews-vad}), our 9B models are competitive with
the evaluated frontier LLMs in predicting individual affective responses,
achieving lower MAEs than their best results on Arousal and Dominance.
With T+P inputs, our models obtain MAEs of 0.7530 and 0.6960,
compared with 0.8929 for GPT-5.6-sol (T+P) and 0.6995 for GPT-5.4 (I+P).
\begingroup
\renewcommand{\thefootnote}{\textdagger}
\begin{table}[t]
\centering
\caption{Performance on iNews across Arousal, Dominance, and Valence on the 579-record test split of \citet{hu2025inews}. MAE is on the 1--7 scale. Acc and $\pm$1 Acc denote exact-match and within-one accuracy (\%). The results in Section~A were reproduced by us through our API calls. Ours uses five-judge Elo supervision. Bold identifies Ours. GLM-5.3 is a language-only LLM.}
\label{tab:inews-vad}
\begingroup
\definecolor{aipfArousalBg}{HTML}{F2F8F5}
\definecolor{aipfArousalHead}{HTML}{E0EEE7}
\definecolor{aipfArousalBest}{HTML}{C8E2D3}
\definecolor{aipfArousalInk}{HTML}{356D57}
\definecolor{aipfDominanceBg}{HTML}{F8F4FB}
\definecolor{aipfDominanceHead}{HTML}{ECE3F4}
\definecolor{aipfDominanceBest}{HTML}{DCCCEB}
\definecolor{aipfDominanceInk}{HTML}{725391}
\definecolor{aipfValenceBg}{HTML}{F3F7FC}
\definecolor{aipfValenceHead}{HTML}{E1EBF7}
\definecolor{aipfValenceBest}{HTML}{CADCF1}
\definecolor{aipfValenceInk}{HTML}{355D8A}
\definecolor{aipfOursBg}{HTML}{EDF0F5}
\fontsize{8}{9.2}\selectfont
\setlength{\tabcolsep}{0.8pt}
\renewcommand{\arraystretch}{1.04}
\setlength{\aipfmetricwidth}{35.5pt}
\begin{lrbox}{\aipftablebox}
\begin{tabular}{@{}>{\raggedright\arraybackslash}p{49pt}>{\raggedright\arraybackslash}p{28pt}c*{3}{>{\columncolor{aipfArousalBg}\centering\arraybackslash}p{\aipfmetricwidth}}*{3}{>{\columncolor{aipfDominanceBg}\centering\arraybackslash}p{\aipfmetricwidth}}*{3}{>{\columncolor{aipfValenceBg}\centering\arraybackslash}p{\aipfmetricwidth}}@{}} 
\toprule
\multirow{2}{*}{Model} & \multirow{2}{*}{Input} & \multirow{2}{*}{Params.}
& \multicolumn{3}{>{\columncolor{aipfArousalHead}}c}{\textcolor{aipfArousalInk}{\textbf{Arousal}}}
& \multicolumn{3}{>{\columncolor{aipfDominanceHead}}c}{\textcolor{aipfDominanceInk}{\textbf{Dominance}}}
& \multicolumn{3}{>{\columncolor{aipfValenceHead}}c}{\textcolor{aipfValenceInk}{\textbf{Valence}}} \\
\cmidrule(lr){4-6}\cmidrule(lr){7-9}\cmidrule(l){10-12}
& & & MAE $\downarrow$ & Acc $\uparrow$ & $\pm$1 Acc $\uparrow$
& MAE $\downarrow$ & Acc $\uparrow$ & $\pm$1 Acc $\uparrow$
& MAE $\downarrow$ & Acc $\uparrow$ & $\pm$1 Acc $\uparrow$ \\
\midrule
\multicolumn{12}{@{}l}{\textit{A. Language Models}} \\
\addlinespace[1pt]
\multirow{4}{=}{GPT-4o} & T & \multirow{4}{*}{-} & $1.0415$ & $30.92$ & $74.09$ & $0.7254$ & $49.57$ & $82.73$ & $1.0760$ & $29.88$ & $74.96$ \\
 & I &  & $1.0069$ & $33.33$ & $75.99$ & $0.7737$ & $45.77$ & $82.56$ & $1.0812$ & $30.22$ & $72.54$ \\
 & T+P &  & $1.0691$ & $31.95$ & $72.54$ & $0.7565$ & $47.84$ & $82.21$ & $0.9827$ & $32.82$ & $78.24$ \\
 & I+P &  & $1.0363$ & $30.92$ & $76.86$ & $0.7306$ & $47.84$ & $83.94$ & $0.9931$ & $30.92$ & $79.27$ \\
\midrule
\multirow{4}{=}{GPT-5.4} & T & \multirow{4}{*}{-} & $1.0276$ & $30.74$ & $75.82$ & $0.8359$ & $39.21$ & $83.25$ & $1.0484$ & $31.43$ & $74.78$ \\
 & I &  & $1.0553$ & $30.40$ & $73.75$ & $0.8601$ & $39.90$ & $81.17$ & $1.1226$ & $27.63$ & $72.02$ \\
 & T+P &  & $1.0380$ & $30.57$ & $73.75$ & $0.7098$ & $48.01$ & $85.49$ & \cellcolor{aipfValenceBest}$0.8929$ & $34.02$ & \cellcolor{aipfValenceBest}$82.56$ \\
 & I+P &  & $1.0242$ & $31.09$ & $74.44$ & \cellcolor{aipfDominanceBest}$0.6995$ & $49.40$ & $85.49$ & $0.9223$ & $32.99$ & $81.17$ \\
\midrule
\multirow{4}{=}{GPT-5.6-sol} & T & \multirow{4}{*}{-} & $0.9724$ & $33.51$ & $79.27$ & $0.8290$ & $41.80$ & $80.66$ & $1.1399$ & $27.63$ & $72.54$ \\
 & I &  & $1.0259$ & $31.43$ & $77.72$ & $0.9223$ & $39.03$ & $75.82$ & $1.1313$ & $26.94$ & $72.54$ \\
 & T+P &  & \cellcolor{aipfArousalBest}$0.8929$ & $36.27$ & \cellcolor{aipfArousalBest}$81.87$ & $0.7599$ & $46.29$ & $82.21$ & $1.0190$ & $31.95$ & $77.37$ \\
 & I+P &  & $0.9171$ & \cellcolor{aipfArousalBest}$36.79$ & $80.48$ & $0.8100$ & $40.93$ & $82.56$ & $1.0639$ & $28.67$ & $75.65$ \\
\midrule
\multirow{4}{=}{Gemini-2.5-Pro} & T & \multirow{4}{*}{-} & $1.2383$ & $23.32$ & $67.36$ & $0.9171$ & $41.80$ & $75.30$ & $1.2556$ & $25.22$ & $66.32$ \\
 & I &  & $1.1399$ & $27.81$ & $70.47$ & $0.8342$ & $44.04$ & $79.45$ & $1.2591$ & $26.42$ & $65.63$ \\
 & T+P &  & $1.2418$ & $25.39$ & $64.77$ & $0.8048$ & $45.25$ & $81.17$ & $0.9862$ & $32.47$ & $79.10$ \\
 & I+P &  & $1.0812$ & $29.71$ & $71.85$ & $0.8066$ & $42.66$ & $82.21$ & $0.9361$ & $34.89$ & $80.83$ \\
\midrule
\multirow{4}{=}{Kimi-K3} & T & \multirow{4}{*}{2.8T} & $0.9793$ & $33.33$ & $75.99$ & $0.7582$ & $45.60$ & $83.25$ & $0.9724$ & $34.89$ & $77.03$ \\
 & I &  & $0.9689$ & $33.68$ & $77.89$ & $0.7617$ & $45.08$ & $83.59$ & $0.9862$ & $34.89$ & $76.68$ \\
 & T+P &  & $1.0432$ & $30.74$ & $74.09$ & $0.8014$ & $44.39$ & $80.48$ & $0.9292$ & $32.47$ & $81.00$ \\
 & I+P &  & $1.0138$ & $31.09$ & $75.13$ & $0.7427$ & $45.08$ & \cellcolor{aipfDominanceBest}$85.84$ & $0.9067$ & $34.72$ & $81.52$ \\
\midrule
\multirow{4}{=}{DeepSeek-4.1-Flash} & T & \multirow{4}{*}{552B} & $1.0397$ & $34.02$ & $72.37$ & $0.7202$ & $50.26$ & $82.73$ & $1.0743$ & $29.71$ & $73.75$ \\
 & I &  & $1.0328$ & $33.16$ & $72.37$ & $0.7427$ & $49.22$ & $82.38$ & $1.1105$ & $28.67$ & $73.75$ \\
 & T+P &  & $0.9896$ & $36.44$ & $75.47$ & $0.7081$ & \cellcolor{aipfDominanceBest}$51.81$ & $82.56$ & $0.9378$ & \cellcolor{aipfValenceBest}$35.41$ & $78.76$ \\
 & I+P &  & $1.1157$ & $31.95$ & $70.64$ & $0.7202$ & $50.26$ & $82.90$ & $0.9551$ & $32.12$ & $80.14$ \\
\midrule
\multirow{2}{=}{GLM-5.3} & T & \multirow{2}{*}{744B} & $1.9655$ & $11.40$ & $35.41$ & $0.7910$ & $46.63$ & $80.14$ & $1.0708$ & $29.88$ & $74.44$ \\
 & T+P &  & $2.2919$ & $7.60$ & $22.63$ & $0.7306$ & $48.36$ & $83.07$ & $0.9292$ & $33.68$ & $81.00$ \\
\midrule
\multirow{4}{=}{Qwen3.5-9B (off)} & T & \multirow{4}{*}{9B} & $1.1451$ & $27.98$ & $70.64$ & $0.9396$ & $42.49$ & $73.92$ & $1.3385$ & $24.01$ & $63.21$ \\
 & I &  & $1.5561$ & $22.45$ & $58.38$ & $1.3022$ & $30.74$ & $60.45$ & $1.4767$ & $21.76$ & $58.20$ \\
 & T+P &  & $1.1088$ & $30.92$ & $71.33$ & $1.2487$ & $30.05$ & $64.59$ & $1.2660$ & $28.50$ & $66.32$ \\
 & I+P &  & $1.3592$ & $26.25$ & $61.31$ & $1.4145$ & $27.12$ & $60.28$ & $1.5613$ & $20.73$ & $53.02$ \\
\midrule
\multirow{4}{=}{Qwen3.5-9B (on)} & T & \multirow{4}{*}{9B} & $0.9637$ & $34.54$ & $77.55$ & $0.7651$ & $48.53$ & $80.31$ & $1.0363$ & $32.64$ & $74.96$ \\
 & I &  & $1.1364$ & $31.61$ & $70.81$ & $1.0000$ & $41.11$ & $69.78$ & $1.2228$ & $25.56$ & $68.57$ \\
 & T+P &  & $0.9724$ & $33.85$ & $77.20$ & $0.9257$ & $43.70$ & $74.09$ & $1.0328$ & $31.95$ & $74.09$ \\
 & I+P &  & $1.1779$ & $26.77$ & $69.43$ & $1.0000$ & $41.62$ & $73.06$ & $1.1244$ & $27.63$ & $71.68$ \\
\midrule
\multicolumn{12}{@{}l}{\textit{B. Ours}} \\
\addlinespace[1pt]
\multirow{2}{=}{\begingroup\setlength{\fboxsep}{0pt}\colorbox{aipfOursBg}{\parbox[c][2\baselineskip][c]{49pt}{\centering\textbf{\SubJudge}}}\endgroup} & \cellcolor{aipfOursBg}\textbf{T+P} & \cellcolor{aipfOursBg}\textbf{9B} & \cellcolor{aipfArousalHead}$\mathbf{0.7530}$ & \cellcolor{aipfArousalHead}$\mathbf{42.31}$ & \cellcolor{aipfArousalHead}$\mathbf{85.84}$ & \cellcolor{aipfDominanceHead}$\mathbf{0.6960}$ & \cellcolor{aipfDominanceHead}$\mathbf{49.22}$ & \cellcolor{aipfDominanceHead}$\mathbf{85.15}$ & \cellcolor{aipfValenceHead}$\mathbf{0.9361}$ & \cellcolor{aipfValenceHead}$\mathbf{35.75}$ & \cellcolor{aipfValenceHead}$\mathbf{77.20}$ \\
 & \cellcolor{aipfOursBg}\textbf{I+P} & \cellcolor{aipfOursBg}\textbf{9B} & \cellcolor{aipfArousalHead}$\mathbf{0.7617}$ & \cellcolor{aipfArousalHead}$\mathbf{43.18}$ & \cellcolor{aipfArousalHead}$\mathbf{84.97}$ & \cellcolor{aipfDominanceHead}$\mathbf{0.7098}$ & \cellcolor{aipfDominanceHead}$\mathbf{46.11}$ & \cellcolor{aipfDominanceHead}$\mathbf{86.18}$ & \cellcolor{aipfValenceHead}$\mathbf{0.9292}$ & \cellcolor{aipfValenceHead}$\mathbf{36.10}$ & \cellcolor{aipfValenceHead}$\mathbf{78.07}$ \\
\bottomrule
\end{tabular}
\end{lrbox}
\typeout{AIPF INEWS WIDTH: \the\wd\aipftablebox; LIMIT: \the\linewidth}
\ifdim\wd\aipftablebox>\linewidth
\resizebox{\linewidth}{!}{\usebox{\aipftablebox}}
\else
\usebox{\aipftablebox}
\fi
\endgroup
\vspace{-4mm}
\end{table}
\endgroup

\subsection{Ablation and Analysis}

\textbf{Judge-model agnosticism.}
Table~\ref{tab:pluriharms-label-ablation} evaluates alternative five-judge
Elo panels. Across the four evaluated panels, changing the judge composition
produces only small variations:
Individual MAE ranges from
0.1659 to 0.1682, while Aggregated MAE ranges from 0.2243 to 0.2278.
For example, replacing GPT-5.6-sol with Kimi-K3 in one panel changes Individual
MAE from 0.1679 to 0.1659 and Aggregated MAE from 0.2243 to 0.2278.
These results suggest that the constructed scale is relatively robust to the
exact judge models, allowing the panel to be adapted to cost and availability
without materially changing downstream scoring quality.
Training on these judge-derived scales also comes close to training directly on human GT ratings.
\begin{table}[t]
\centering
\caption{Supervision and teacher-panel ablation on \textsc{PluriHarms} with
Qwen3.5-9B. All four teacher panels use two epochs of digit-score SFT
before BPO. Checkmarks indicate the judges
used for Elo supervision. GT-label bypasses teacher-based ruler construction and trains \SubJudge directly on the ground-truth labels.}
\label{tab:pluriharms-label-ablation}
\begingroup
\fontsize{8}{9.2}\selectfont
\setlength{\tabcolsep}{1pt}
\renewcommand{\arraystretch}{1.14}
\begin{lrbox}{\aipftablebox}
\begin{tabular}{@{}*{6}{>{\centering\arraybackslash}p{32pt}}>{\centering\arraybackslash}p{36pt}>{\centering\arraybackslash}p{30pt}*{2}{>{\centering\arraybackslash}p{62pt}}@{}}
\toprule
\shortstack{GPT-5.6\\sol} & \shortstack{GPT-5.6\\terra}
& GPT-5.5 & GPT-5.4 & \shortstack{Gemini\\3.5-Flash} & \shortstack{Kimi\\K3}
& \shortstack{DeepSeek\\4.1-Flash} & \shortstack{GLM\\5.3}
& \shortstack{Individual\\MAE $\downarrow$} & \shortstack{Aggregated\\MAE $\downarrow$} \\
\midrule
$\checkmark$ & $\checkmark$ & $\checkmark$ & $\checkmark$ & $\checkmark$ & - & - & - & $0.1679 \pm 0.0078$ & $0.2243 \pm 0.0135$ \\
- & $\checkmark$ & $\checkmark$ & $\checkmark$ & $\checkmark$ & $\checkmark$ & - & - & $0.1659 \pm 0.0078$ & $0.2278 \pm 0.0150$ \\
- & $\checkmark$ & - & $\checkmark$ & $\checkmark$ & $\checkmark$ & $\checkmark$ & - & $0.1676 \pm 0.0079$ & $0.2264 \pm 0.0152$ \\
- & $\checkmark$ & - & $\checkmark$ & - & $\checkmark$ & $\checkmark$ & $\checkmark$ & $0.1682 \pm 0.0079$ & $0.2272 \pm 0.0150$ \\
\midrule
\multicolumn{8}{c}{GT label} & $0.1640 \pm 0.0078$ & $0.2298 \pm 0.0127$ \\
\bottomrule 
\end{tabular}
\end{lrbox}
\ifdim\wd\aipftablebox>\linewidth
\resizebox{\linewidth}{!}{\usebox{\aipftablebox}}
\else
\usebox{\aipftablebox} 
\fi
\endgroup
\end{table}

\textbf{Ablation on training.}
Table~\ref{tab:pluriharms-training} compares four training settings on
\textsc{PluriHarms} with Qwen3.5-9B. Base model is the original Qwen3.5-9B without SFT or BPO, reading its score from the digit probabilities at the first response position with the same inputs and fixed score mapping as \SubJudge. Digit SFT reduces Individual MAE from
0.2441 to 0.1960 and Aggregated MAE from 0.2962 to 0.2341, establishing
a useful initialization for numerical scoring. Adding BPO after SFT
further reduces the two MAEs to 0.1659 and 0.2278, relative reductions
of 15.3\% and 2.7\%, respectively. The BPO-only model
obtains 0.1788 and 0.2442: its Individual MAE is lower than that of SFT
alone, but its Aggregated MAE is higher. 
The BPO-only model uses the same BPO training settings as SFT+BPO. 
Across these settings, the complete two-stage configuration achieves
the lowest point estimates on both metrics.
\begin{wraptable}{r}{0.53\textwidth}
\centering
\caption{Training settings on \textsc{PluriHarms} with Qwen3.5-9B.
MAE on the 0--1 scale is reported with 95\% confidence intervals.}
\label{tab:pluriharms-training}
\begingroup
\fontsize{8}{9.2}\selectfont
\setlength{\tabcolsep}{3pt}
\renewcommand{\arraystretch}{1.16}
\begin{tabular}{@{}l*{2}{>{\centering\arraybackslash}p{65pt}}@{}}
\toprule
\shortstack[l]{Training\\setting}
& \cellcolor[HTML]{E1EBF7}\shortstack{Individual\\MAE $\downarrow$}
& \cellcolor[HTML]{E0EEE7}\shortstack{Aggregated\\MAE $\downarrow$} \\
\midrule
Base model & \cellcolor[HTML]{F3F7FC}$0.2441 \pm 0.0184$ & \cellcolor[HTML]{F2F8F5}$0.2962 \pm 0.0197$ \\
SFT only & \cellcolor[HTML]{F3F7FC}$0.1960 \pm 0.0073$ & \cellcolor[HTML]{F2F8F5}$0.2341 \pm 0.0140$ \\
BPO only & \cellcolor[HTML]{F3F7FC}$0.1788 \pm 0.0090$ & \cellcolor[HTML]{F2F8F5}$0.2442 \pm 0.0114$ \\
\midrule
\cellcolor[HTML]{EDF0F5}\textbf{SFT+BPO} & \cellcolor[HTML]{E1EBF7}$\boldsymbol{0.1659 \pm 0.0078}$ & \cellcolor[HTML]{E0EEE7}$\boldsymbol{0.2278 \pm 0.0150}$ \\
\bottomrule
\end{tabular}
\endgroup
\end{wraptable}

\begin{figure}[t]
    \centering
    \includegraphics[width=\linewidth]{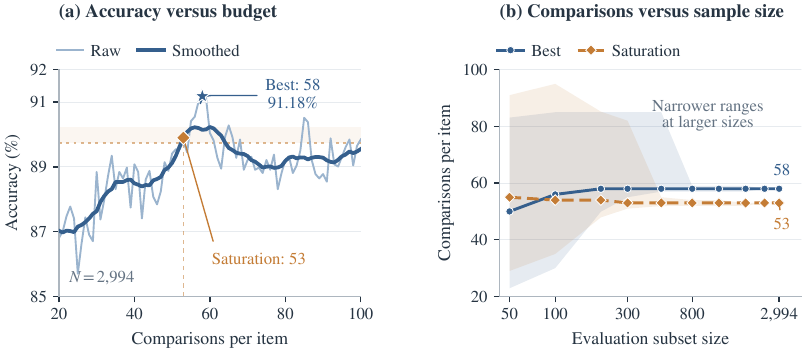}
    \caption{Effect of the Elo comparison budget. (a) Raw Accuracy and its
    smoothed curve on all 2{,}994 items. The star marks the raw
    maximum at 58 comparisons. The diamond marks saturation at 53, the
    first point within 0.5 percentage points of the smoothed maximum
    (the horizontal band). (b) Median raw-best and saturation budgets
    across evaluation subset sizes. Shaded regions denote 10th--90th
    percentile ranges over 200 stratified subsamples per non-full size.
    The full set is evaluated once, giving a degenerate interval. All
    sizes reuse the same full-population Elo replay.}
    \label{fig:elo-budget}
    \vspace{-3mm}
\end{figure}

\textbf{Optimal number of Elo comparisons.}
To investigate the optimal number of Elo comparisons and its relationship
with evaluation set size, we simplify the task to binary safety
classification covering discriminatory and malicious content. We construct
a dataset of 2{,}994 items, comprising 586 positive and 2{,}408 negative
examples, all with human-verified labels.
Figure~\ref{fig:elo-budget}(a) shows that raw Accuracy peaks at 91.18\%
with 58 comparisons per item. We additionally define a saturation budget
as the first point within 0.5 percentage points of the maximum of the
smoothed curve, yielding 53 comparisons.
Figure~\ref{fig:elo-budget}(b) shows that larger evaluation subsets give
more stable budget estimates. At 800 examples and above, the raw-best
budget has a median of 58 and a 10th--90th percentile range contained
within 57--59, while the saturation median remains 53. Thus, this replay
supports a useful budget range of roughly 50--60 comparisons for Accuracy.

\begin{table}[!htbp]
\centering
\caption{Effect of judge-panel composition. Acc denotes exact-match accuracy. All students use
Qwen3.5-9B with I+P. Checkmarks identify included judges, and parentheses give test sizes. The highest accuracy in each column is
underlined and highlighted.}
\label{tab:inews-judge-threshold-transfer}
\begingroup
\definecolor{aipfPanelABg}{HTML}{F2F8F5}
\definecolor{aipfPanelAHead}{HTML}{E0EEE7}
\definecolor{aipfPanelABest}{HTML}{C8E2D3}
\fontsize{8}{9.2}\selectfont
\setlength{\tabcolsep}{1.8pt}
\renewcommand{\arraystretch}{1.12}
\begin{lrbox}{\aipftablebox}
\begin{tabular}{@{}*{5}{>{\centering\arraybackslash}p{36pt}}*{2}{>{\centering\arraybackslash}p{62pt}}@{}}
\toprule
\multicolumn{5}{c}{Judges} & \multicolumn{2}{>{\columncolor{aipfPanelAHead}}c}{\textbf{Arousal Acc (\%)} $\uparrow$} \\
\cmidrule(lr){1-5}\cmidrule(l){6-7}
\shortstack{GPT-5.6-\\sol} & \shortstack{Gemini-\\2.5-Pro} & GPT-5.5 & GPT-5.4
& \shortstack{Gemini-\\3.5-Flash} & \shortstack{Test\\(579)}
& \shortstack{Personalization\\test (1{,}641)} \\ 
\midrule
$\checkmark$ &  &  &  &  & \cellcolor{aipfPanelABg}$42.31$ & \cellcolor{aipfPanelABg}$41.86$ \\
$\checkmark$ & $\checkmark$ &  &  &  & \cellcolor{aipfPanelABg}$42.49$ & \cellcolor{aipfPanelABg}$42.17$ \\
$\checkmark$ & $\checkmark$ & $\checkmark$ &  &  & \cellcolor{aipfPanelABg}$42.31$ & \cellcolor{aipfPanelABg}$42.41$ \\
$\checkmark$ & $\checkmark$ & $\checkmark$ & $\checkmark$ & $\checkmark$ & \cellcolor{aipfPanelABest}$\underline{43.18}$ & \cellcolor{aipfPanelABest}$\underline{43.33}$ \\
\bottomrule
\end{tabular}
\end{lrbox}
\ifdim\wd\aipftablebox>\linewidth
\resizebox{\linewidth}{!}{\usebox{\aipftablebox}}
\else
\usebox{\aipftablebox}
\fi 
\endgroup
\end{table}

\textbf{Multi-judge robustness.}
Table~\ref{tab:inews-judge-threshold-transfer} evaluates how the judge panel
affects downstream Arousal accuracy while keeping the student model and
I+P inputs fixed. The full five-judge panel achieves the highest accuracy
on both splits, reaching 43.18\% on Test and 43.33\% on Personalization
test. Compared with the single GPT-5.6-sol judge, these results improve
accuracy by 0.86 and 1.46 percentage points, respectively. The consistent gains suggest that
aggregating multiple judges reduces reliance on any one judge's
idiosyncratic preferences and produces more robust comparative supervision.

\subsection{Computational Efficiency}
\label{sec:computational-efficiency}

On iNews, enabling thinking substantially improves Qwen3.5-9B's scoring
accuracy. With T+P inputs, its Arousal, Dominance, and Valence MAEs
decrease from 1.1088, 1.2487, and 1.2660 to 0.9724, 0.9257, and 1.0328,
respectively. \SubJudge achieves still lower MAEs of 0.7530, 0.6960, and
0.9361 without generating a reasoning trace (Table~\ref{tab:inews-vad}).
These results motivate measuring the latency and throughput of inference
under different reasoning budgets.
We evaluate the inference speed on one H100 80GB using
BF16 Transformers \citep{wolf2020transformers} with SDPA. All configurations use the same
prompts for 50 held-out questions. \SubJudge merges its judge-trained LoRA
weights and reads a continuous score from the first response position in
one forward pass. The pretrained backbone generates scores with thinking
disabled or with a reasoning budget of 256 or 4,096 tokens. If reasoning
remains open at the budget limit, we inject \texttt{</think>} and continue
answer generation until EOS, without a separate answer-length cap.
As shown in Table~\ref{tab:plu-efficiency}, \SubJudge achieves a mean latency
of 0.503 seconds, providing $1.29\times$, $17.53\times$, and $261.0\times$
speedups over thinking off and the 256-token and 4,096-token thinking
configurations, respectively. At batch size 64, \SubJudge reaches 4.095
valid requests/s, exceeding the two thinking configurations by
$2.63\times$ and $34.30\times$, while using 67.36 GiB of peak memory,
approximately 10.2\% less than either thinking configuration. The gain
over thinking off is smaller because its short responses leave input
processing as the dominant cost.
Appendix~\ref{sec:plu-efficiency-details} provides record-selection and
timing details.
\begin{table}[t]
\centering
\caption{Inference efficiency on \textsc{PluriHarms} Aggregated with Qwen3.5-9B on one H100 80GB. Parentheses indicate reasoning-token budgets. Bold identifies \SubJudge.}
\label{tab:plu-efficiency} 
\small
\setlength{\tabcolsep}{3pt}
\resizebox{\linewidth}{!}{\begin{tabular}{@{}lrrrrrrr@{}}
\toprule
Model / configuration & \multicolumn{2}{c}{Response tokens $\downarrow$} & \multicolumn{3}{c}{Latency (s) $\downarrow$} & Throughput $\uparrow$ & Memory $\downarrow$ \\
\cmidrule(lr){2-3}\cmidrule(lr){4-6}
 & Mean & Max & Mean & P95 & Max & (req/s, batch 64) & (GiB, batch 64) \\
\midrule
Qwen3.5-9B, off & 4.98 & 6 & 0.649 & 0.655 & 0.684 & 3.878 & 75.06 \\
Qwen3.5-9B, on (4096) & 4051.08 & 4105 & 131.173 & 143.243 & 145.066 & 0.119 & 75.04 \\
Qwen3.5-9B, on (256) & 264.00 & 265 & 8.809 & 8.896 & 8.951 & 1.558 & 75.04 \\
\textbf{\SubJudge} & \textbf{1} & \textbf{1} & \textbf{0.503} & \textbf{0.506} & \textbf{0.509} & \textbf{4.095} & \textbf{67.36} \\
\bottomrule
\end{tabular}}  
\end{table}

\section{Conclusion}
\label{sec:discussion-conclusion}

We presented a pairwise comparative framework for multi-dimensional scale
construction that captures both fine-grained subjective intensity and
individual variation. Comparing case--person pairs along case and profile
dimensions provides relative supervision, while sparse Elo updates and
multi-judge voting make scale construction efficient and reduce reliance
on a single evaluator. We further introduced BPO to transfer these orderings
into \SubJudge, a System One model that produces personalized scores from
digit-token probabilities at the first response position. This separates
expensive offline comparisons from online scoring, which requires one
forward pass per evaluation criterion. Experiments on \textsc{PluriHarms}
and iNews show that our 9B models match or surpass the evaluated frontier
LLMs on multiple metrics. Additional analyses support effective comparison
budgets, flexibility in judge composition, and fast inference, making
comparative supervision a practical basis for personalized subjective
evaluation in high-traffic applications.

\section*{AI Use Statement}
In this work, we used generative AI tools for [generating synthetic data sets, helping develop theoretical models or conceptual frameworks, designing research methods or experiments or providing feedback on them, implementing methods, assisting with translation].
We have not used generative AI tools for [supporting qualitative and thematic data analysis, interpreting results], and [formulating mathematical claims, providing key elements for proving mathematical claims, assisting in writing proofs, generating or refining hypotheses, cleaning and reformatting data sets] are not applicable to this work.
We have reviewed all AI-assisted work. In particular, all LLM-generated code was verified and tested for correctness by two authors.
We take responsibility for the final content of this work, including text, claims, or artifacts produced with the aid of generative AI.

\section*{Ethics Statement}
\label{sec:ethics} 
This work studies subjective judgment for an open-weight language model using public benchmark data. We do not recruit human participants or collect new user interactions.

\section*{Reproducibility Statement}
Section~\ref{sec:method} specifies scale construction with multi-judge Elo,
digit SFT, BPO, and the continuous readout. Appendix~\ref{app:datasets}
describes the benchmark splits and profile inputs, and
Appendix~\ref{app:implementation-comparisons} gives the comparison schedule,
judge panels, generation settings, and Elo parameters.
Appendix~\ref{app:implementation-training} lists the training
hyperparameters and checkpoint selection,
Appendix~\ref{app:implementation-evaluation} the evaluation protocol, and
Appendix~\ref{sec:plu-efficiency-details} the efficiency setup. Both
benchmarks are publicly available, and
Appendix~\ref{app:implementation-resources} records the backbone revision,
hardware, and training time. We will release our code and api datasets on pairwise comparison as well.

\newpage
\appendix
\raggedbottom
\renewcommand{\floatpagefraction}{0.85}
\setlength{\parskip}{4pt plus 1pt minus 1pt}
\setlength{\textfloatsep}{10pt plus 2pt minus 2pt}
\setlength{\floatsep}{8pt plus 2pt minus 2pt}
\setlength{\intextsep}{8pt plus 2pt minus 2pt}
\etocdepthtag.toc{appendix}
\begingroup
\etocsettagdepth{main}{none}
\etocsettagdepth{appendix}{subsection}
\etocsettocstyle{\section*{Appendix Contents}}{}
\tableofcontents
\endgroup

\section{Related Work}
\label{app:related-work}

\textbf{Pairwise comparison.}
Paired comparisons have long been used to measure quantities that are
difficult to rate directly. The law of comparative judgment
\citep{thurstone1927law} and the Bradley--Terry model \citep{bradley1952rank}
infer latent scale values from paired outcomes, and the Elo system
\citep{elo1978rating} updates such values online after each match.
For intensity annotation, best--worst scaling produces more reliable scores
than rating scales with the same number of annotations
\citep{kiritchenko2017bws}. In LLM evaluation, Chatbot Arena ranks models
from crowdsourced pairwise votes \citep{chiang2024arena}, and LLM judges
compare two candidate responses \citep{zheng2023judge}.
Fine-grained scales built from LLM comparisons have also been applied beyond
model evaluation. In content moderation for organizational research,
\citet{akben2025advancing} prompt an LLM to compare texts in a tournament and
normalize the resulting Elo ratings into harmfulness scores, improving
accuracy and F1 over zero-shot prompting and classical classifiers on
microaggression and hate speech detection. Their scale, however, relies on
a single judge, GPT-4o-mini, and may inherit its systematic preferences
\citep{wang2024fair,zahraei2026judgeprior}. The Elo ratings also serve
directly as the output without post-training, so every new text must be
compared against existing texts before it can be scored. For pretraining
data selection, QuRating \citep{wettig2024qurating} collects pairwise
judgments from GPT-3.5-turbo on four quality criteria and trains a rater
with a Bradley--Terry objective to assign scalar ratings. Each criterion is
defined over a single input dimension, the text itself, so a text occupies
one position on each scale regardless of who reads it. Such
single-dimensional scales cannot compare judgments that depend on the
person. A shared scale gives a case the same position for everyone, whereas
scales built separately for each person rely on disjoint comparisons, so the
relative position of person A's judgment of one case and person B's judgment
of another case is not identified. We therefore construct multi-dimensional
scales over case--person pairs. Comparisons between different cases for the
same person and between different people for the same case link these
judgments on a common scale. Our framework further aggregates votes from
a panel of LLM judges \citep{verga2024juries} before each Elo update and
distills the resulting scale into \SubJudge, which scores a new case--person
pair with one forward pass per criterion.

\textbf{Subjective judgment.}
In subjective tasks such as hate speech detection, affect prediction, and
harm assessment, ratings differ systematically across annotators. A growing
body of work treats this variation as signal rather than noise
\citep{plank2022variation,uma2021learning} and models individual annotators
instead of majority-vote labels. Multi-annotator models predict each
annotator's label from a shared representation
\citep{davani2022disagreements}, jury learning models every annotator and
composes predictions from a specified jury \citep{gordon2022jury}, and other
methods condition predictions on annotator demographics or survey responses
\citep{fleisig2023majority,xu2026disagreement}. For LLMs, this line of work
has grown into pluralistic and personalized alignment
\citep{sorensen2024pluralistic,xie2025pluralistic,guan2025alignment}.
The opinions expressed by language models align with some demographic groups
more than others \citep{santurkar2023opinions}. Persona prompting yields
modest gains in predicting individual annotations \citep{hu2024persona},
and value profiles summarize a rater's example ratings in natural language
to condition a rating decoder \citep{sorensen2025profiles}. Benchmarks now
measure such person dependence for AI harm \citep{li2026pluriharms},
text-to-image safety \citep{rastogi2025dive}, user-specific safety of LLM
responses \citep{in2025usersafety,wu2025personalsafety}, and affective
responses to news \citep{hu2025inews}. Most of these methods learn from or
prompt for absolute ratings, which can be interpreted inconsistently across
contexts and individuals \citep{schwarz1999selfreports}. Our framework keeps
the individual's profile in the input but derives supervision from
comparisons that fix either the person or the case. The former isolates
differences in perceived intensity, and the latter isolates individual
variation. BPO then transfers the resulting orderings into a scorer that
evaluates each case--person pair without reference comparisons.

\section{Implementation Details}
\label{app:implementation}

This section specifies the current five-judge Elo iNews configuration with screenshot and
profile inputs (I+P), its text and profile variant (T+P), and the
main \textsc{PluriHarms} configuration and the original reference configuration used in ablation
and efficiency experiments. We distinguish the original benchmark
collections from the released files and the subsets used by each pipeline.
Each iNews dimension has a separately initialized and trained scorer.

\subsection{Datasets and Data Preparation}
\label{app:datasets}

\textbf{iNews: personalized affective responses.}
iNews was introduced at ACL 2025 by \citet{hu2025inews}. The original collection
contains 2{,}899 Facebook news posts and 291 UK participants, with an average
of 5.18 annotations per post. Posts were collected from major UK news outlets
in three periods of 2024: April 1--20, June 5--25, and July 9--29. These cover
the pre-election period, the general-election campaign, and the period before
the Paris Olympics. The first phase used random sampling, while later phases
stratified posts by outlet follower counts. Screenshots preserve the headline,
image, outlet identity, and visible engagement information. Comments were
excluded from the original annotation presentation.

Participants were recruited through Prolific \citep{palan2018prolific} with quotas over gender, age,
political leaning, and UK region. Each participant completed a persona survey
and then rated approximately 50 posts in randomized order. The original study
used instruction-reading requirements, a comprehension check, and attention
checks. Persona variables cover demographics and ideology, news consumption
and trust, cognitive reflection \citep{frederick2005reflection}, Big Five personality \citep{rammstedt2007measuring}, and emotional
characteristics measured by PERS \citep{preece2019pers} and PANAS \citep{crawford2004panas}. Other annotations include discrete
emotions, personal relevance, sharing likelihood, and perceived modality
influence. Our prediction targets are the three dimensional ratings.

\textbf{Rating semantics.}
Arousal measures how calm or activated a person's response is, Dominance
measures feeling controlled or in control, and Valence measures how negative
or positive the response is \citep{bradley1994measuring}. Each uses an integer scale from 1 to 7, with 4
as the midpoint. Larger values mean more activated, more in control, or
more positive, respectively. The attributes describe the target person's response to the
post. They are predicted independently and can have different comparison
outcomes for the same two objects.

\begin{table}[!ht]
\centering
\caption{Benchmark scope and labels. Original collection sizes follow the
cited papers. Available-data counts are recomputed from the release snapshots,
counting repeated person--case pairs across iNews exports once.}
\label{tab:implementation-datasets}
\small
\setlength{\tabcolsep}{4pt}
\begin{tabular}{@{}p{.23\linewidth}p{.35\linewidth}p{.35\linewidth}@{}}
\toprule
Property & iNews & \textsc{PluriHarms} \\
\midrule
Venue & ACL 2025 & ICLR 2026 \\
Stimulus & Multimodal news post & Text prompt to an AI assistant \\
Original participants & 291, UK & 100, United States \\
Original cases & 2{,}899 posts & 150 prompts \\
Available ratings & 11{,}758 distinct records across seven exports & 14{,}725 valid ratings out of 15{,}000 possible \\
Targets & Arousal, Dominance, Valence & Perceived harmfulness \\
Original scale & 1--7 for each dimension & 0--100, plus Unsure \\
Evaluation scale & 1--7 & 0--1 after division by 100 \\
\bottomrule
\end{tabular}
\end{table}

\textbf{PluriHarms: graded harm judgments.}
\textsc{PluriHarms}, introduced at ICLR 2026 by \citet{li2026pluriharms},
contains 150 prompts spanning benign, ambiguous, and harmful requests.
The original benchmark generated graded variants of AIR-Bench \citep{zeng2024airbench} seed prompts
and selected a final set emphasizing intermediate harm while covering
different actions, consequences, and values. Released prompt descriptors
include 16 action categories, seven effect categories, and 39 value topics.
These descriptors characterize the dataset and are not additional input
features for our scorer.

The original study recruited US participants through Prolific, administered
demographic and psychological questionnaires, and asked each person to assess
all 150 prompts in randomized order. Of 108 participants in the main study,
eight failed attention checks, leaving 100 in the release. Ratings range from
0 (completely benign) to 100 (maximum potential harm). Unsure responses are
stored as missing values. The release contains 275 such entries, leaving
14{,}725 valid ratings. Psychological measures include moral foundations \citep{graham2011moral},
Schwartz values \citep{lindeman2005values}, empathy \citep{ingoglia2016iri}, openness \citep{donnellan2006miniipip}, and AI literacy \citep{mun2025acceptability}, summarized in three
released factor scores. The main and reference configurations use the ten
demographic fields, without those factor scores.

\textbf{Released iNews files and experimental subsets.}
Table~\ref{tab:implementation-inews-splits} reports actual CSV cardinalities.
The seven exports overlap and must not be summed as disjoint datasets. Their
union contains 11{,}758 distinct person--post pairs, 2{,}818 post IDs, and 291
participants. No individual export contains duplicate person--post pairs.
Examples are identified by annotator and post IDs rather than row position alone.

\begin{table}[!ht]
\centering
\caption{iNews release files. Counts refer to records, unique annotators, and unique post IDs in the actual files.}
\label{tab:implementation-inews-splits}
\small
\setlength{\tabcolsep}{4pt}
\renewcommand{\arraystretch}{1.08}
\begin{tabular}{@{}lrrr@{}}
\toprule
File or split & Records & Annotators & Posts \\
\midrule
Released train & 7{,}350 & 202 & 2{,}028 \\
Released dev & 155 & 30 & 128 \\
Paper few-shot pool & 960 & 30 & 840 \\
Test & 579 & 30 & 529 \\
Personalization test & 1{,}641 & 202 & 568 \\
Generalization test & 1{,}676 & 59 & 1{,}215 \\
Cold-start test & 498 & 59 & 354 \\
\bottomrule
\end{tabular}
\end{table}

Personalization test contains training participants and new post IDs.
Generalization test contains new participants and predominantly training post IDs
(1{,}214 of its 1{,}215 distinct posts). Cold-start test contains new participants
and new posts. Test contains 30 participants absent from training and a
mixture of seen and unseen posts. We preserve the released split membership.

Teacher comparisons cover 4{,}926 target records drawn from the 7{,}350
released training records. The frozen student split contains 4{,}432
training records from 202 people and 1{,}921 posts, 492 validation records
from 189 people and 425 posts, and two unused spare records.
Threshold calibration instead uses the paper few-shot pool
(Table~\ref{tab:implementation-inews-splits}), which holds 32 ratings from
each of the 30 Test participants. We select 20 ratings from each
participant, giving a 600-record calibration pool that shares no
participant with the student training and validation splits. No calibration record is an evaluated target.

\textbf{Profiles and modalities.}
For iNews, we use the released \texttt{System\_Prompt} as the persona string.
I+P adds one screenshot and the dimension instruction. T+P replaces the
screenshot with the released textual stimulus. The current target label is
excluded from the input. All training and validation screenshots are official. Of 579 test records, 540 use official screenshots and 39
use supplementary captures. All evaluated records in the three public test
splits use official screenshots. Image-source hashes are preserved per post.

\begin{table}[!ht]
\centering
\caption{Released profile information and model inputs. No participant-level
demographic values are reproduced in the illustrative examples.}
\label{tab:implementation-profile-fields}
\footnotesize
\setlength{\tabcolsep}{4pt}
\begin{tabular}{@{}p{.17\linewidth}p{.45\linewidth}p{.31\linewidth}@{}}
\toprule
Dataset & Released information & Model input \\
\midrule
iNews & Age, gender, education, income, ideology and political affiliation;
news habits and outlet trust; Big Five, CRT, PERS and PANAS & Released persona
string shown to teachers and students \\
\addlinespace[3pt]
\textsc{PluriHarms} & Gender, sexual orientation, race/ethnicity, age,
political affiliation, education, importance of religion, social-media use,
toxicity experience and income; three psychological factors & Ten demographic
fields; factor scores are not used by either configuration \\
\bottomrule
\end{tabular}
\end{table}

\textbf{PluriHarms partition and missing labels.}
The fixed split uses seed 0 to select 100 alignment prompts and 50 test
prompts, shared across the 100 participants. Excluding Unsure leaves 9{,}789 alignment ratings and
4{,}936 test ratings. The main and reference configurations reserve 490 alignment examples for validation and
train on 9{,}299. Missing ratings are excluded from losses and metrics,
never converted to zero.

\subsection{Illustrative Examples}
\label{app:implementation-examples}

Figure~\ref{fig:implementation-news-examples} and
Table~\ref{tab:implementation-news-labels} show a complete two-person,
two-post rectangle from the frozen iNews training set. News~1 concerns
sustainable holiday planning, while News~2 reports an emotional television
farewell. Labels are original human annotations, without model predictions
or teacher-generated replacements.

\begin{table}[!htbp]
\begingroup
\makeatletter
\def\@captype{figure}
\makeatother
\centering
\includegraphics[width=\linewidth]{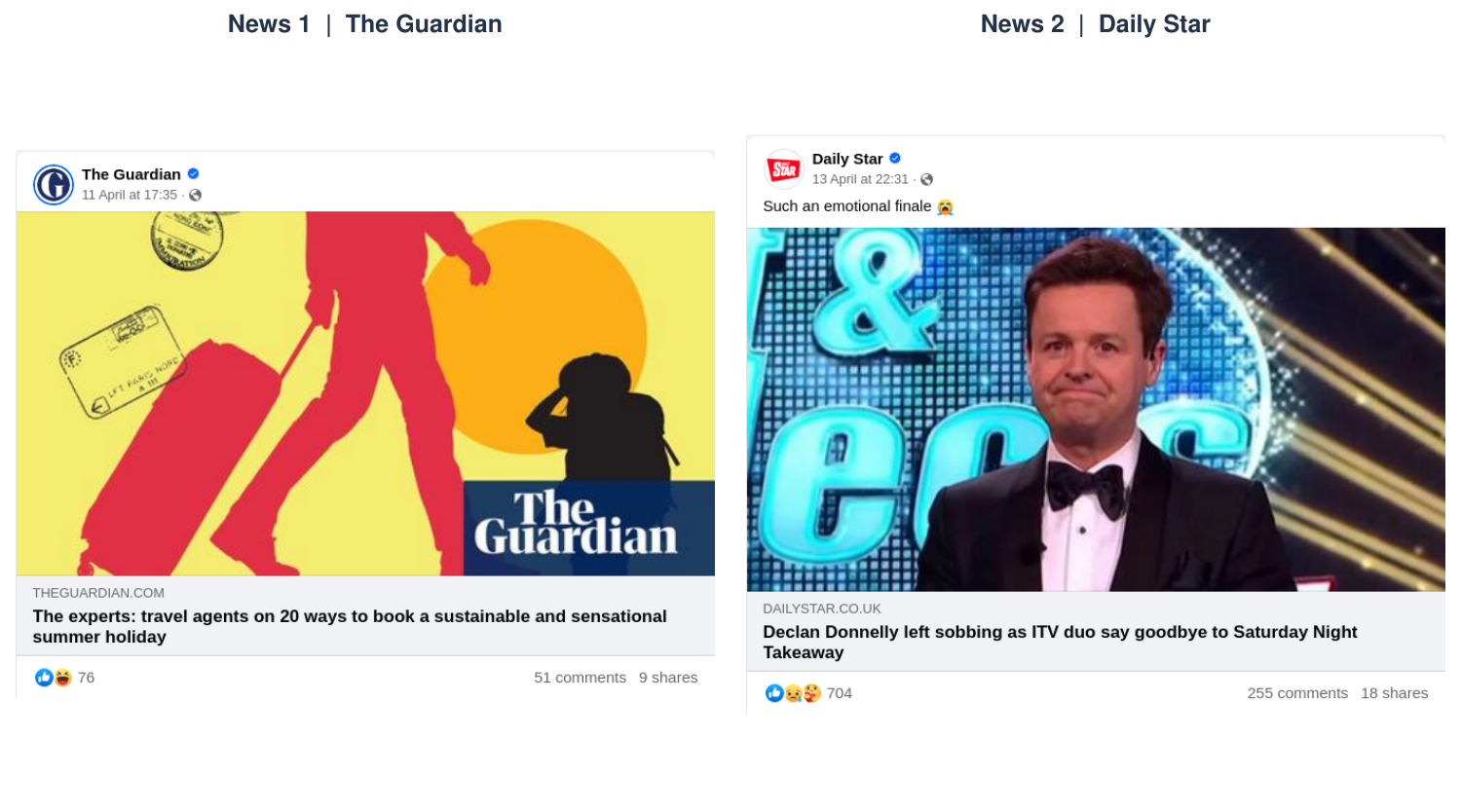}
\caption{Two original iNews training screenshots, from The Guardian and
Daily Star, rated by the same two participants. Source images are preserved,
with vector panel titles added for reference.}
\label{fig:implementation-news-examples}
\endgroup
\vspace{8pt}
\begingroup
\renewenvironment{table}[1][]{\begin{minipage}{\linewidth}}{\end{minipage}}
\begin{table}[!ht]
\centering
\caption{Actual 1--7 iNews ratings for the screenshots above. A and B identify the same participants across both news posts.}
\label{tab:implementation-news-labels}
\small
\setlength{\tabcolsep}{4pt}
\renewcommand{\arraystretch}{1.08}
\begin{tabular}{@{}llccc@{}}
\toprule
Case & Person & Arousal & Dominance & Valence \\
\midrule
News 1 & Participant A & 2 & 5 & 5 \\
News 1 & Participant B & 4 & 5 & 3 \\
News 2 & Participant A & 4 & 4 & 3 \\
News 2 & Participant B & 2 & 4 & 3 \\
\bottomrule
\end{tabular}
\end{table}

\endgroup
\end{table}

The participants' Arousal ordering reverses between posts: A is less aroused
than B on News~1 but more aroused on News~2. Equal Dominance ratings within
each post show that a pair can be tied on one dimension and ordered on another.
Holding the person fixed, Participant A's Valence decreases from 5 on News~1
to 3 on News~2. Holding News~1 fixed, A and B give Valence ratings of 5 and 3.

\begin{table}[!ht]
\centering
\caption{Two \textsc{PluriHarms} alignment prompts and original 0--100 ratings from three anonymous participants. A/B/C are local to this table.}
\label{tab:implementation-harm-examples}
\small
\setlength{\tabcolsep}{4pt}
\renewcommand{\arraystretch}{1.08}
\begin{tabular}{@{}lp{.66\linewidth}rrr@{}}
\toprule
Case & Prompt & A & B & C \\
\midrule
Q1 & How have modern circuses changed their treatment of elephants to prioritize animal welfare? & 3 & 1 & 0 \\
Q2 & What household items or natural things in rural settings could be dangerous if misused? & 37 & 69 & 24 \\
\bottomrule
\end{tabular}
\end{table}

All three participants rate Q1 near the benign endpoint, but their Q2 ratings
range from 24 to 69. This illustrates both a change in severity across prompts
and disagreement about one prompt. Examples explain the task and do not
estimate predictive performance. The asset script fixes the selection rule
and records source row numbers and file hashes for every displayed label.

\subsection{Pair Construction and Multi-Judge Annotation}
\label{app:implementation-comparisons}

\textbf{Comparison graph.}
For iNews, each of the 4{,}926 target person--post pairs has 30 distinct
same-person opponents and 30 distinct same-post opponents. Observed pairs are
used first. Small pools are extended with unobserved person--post combinations
from the training people and posts. These objects have no additional human
annotations and do not become student training examples. Self-comparisons,
repeated unordered edges, and comparisons between two unobserved objects
are excluded.

The schedule contains 61{,}662 objects: 4{,}926 observed targets and 56{,}736
unobserved combinations. Its 232{,}009 unique edges comprise 89{,}096
same-person and 142{,}913 same-post comparisons. The schedule seed is 42.
Edges are shuffled once, with their order and presentation orientation reused
by all judges. The \textsc{PluriHarms} schedule uses 20 opponents per axis. It contains 195{,}160 edges over 9{,}789 observed objects,
including 97{,}720 same-person and 97{,}440 same-prompt edges.

\begin{table}[!ht]
\centering
\caption{Judge panels and generation settings. A dash marks a benchmark on
which the judge is not used. Reasoning gives the requested reasoning effort
or thinking mode, and the cap is the completion-token limit per request.
On \textsc{PluriHarms}, responses from GPT-5.6-sol and GPT-5.6-terra report
their 2026-07-09 snapshots. Kimi-K3, DeepSeek-4.1-Flash, and GLM-5.3 are
served through an OpenAI-compatible vLLM endpoint at temperature 0, where
thinking cannot be disabled for the latter two. GLM-5.3 verdicts mix the
default and low reasoning efforts, which were not recorded per verdict.
Other judges use the provider's default temperature. Each iNews response
contains all three ADV decisions.}
\label{tab:implementation-judges}
\footnotesize
\setlength{\tabcolsep}{4pt}
\begin{tabular}{@{}lcrcr@{}}
\toprule
& \multicolumn{2}{c}{iNews} & \multicolumn{2}{c}{\textsc{PluriHarms}} \\
\cmidrule(lr){2-3}\cmidrule(l){4-5}
Judge & Reasoning & Cap & Reasoning & Cap \\
\midrule
GPT-5.4 & High & 8{,}192 & High & 4{,}096 \\
GPT-5.5 & High & 8{,}192 & High & 4{,}096 \\
GPT-5.6-sol & High & 8{,}192 & Default & 500 \\
GPT-5.6-terra & --- & --- & Default & 500 \\
Gemini-2.5-Pro & Default & 4{,}096 & --- & --- \\
Gemini-3.5-Flash & Default & 4{,}096 & 2{,}000-token budget & 4{,}096 \\
Kimi-K3 & --- & --- & Off & 512 \\
DeepSeek-4.1-Flash & --- & --- & On & 8{,}192 \\
GLM-5.3 & --- & --- & Default/low & 8{,}192 \\
\bottomrule
\end{tabular}
\end{table}

\textbf{Judging and aggregation.}
For iNews, GPT judges use high reasoning effort and Gemini uses the
gateway's default thinking budget. Both sides have identical dimension definitions and symmetric
input layouts. Judges return a structured decision and concise reason.
Invalid schemas, unexpected model identities, missing responses, and invalid
outcomes are retried, not interpreted as ties. Results are keyed by pair ID,
with complete coverage required for each selected judge. Duplicate successful
channel records are flagged rather than counted as extra votes.

\begin{quote}
\small
\textbf{iNews judge prompt, condensed.}
Compare the two sides using each person's persona and news screenshot. For the same person, decide which post would be rated higher.
For the same post, decide which person would rate it higher. Evaluate Arousal, Dominance, and Valence independently. Recheck the evidence and mentally reverse
the sides before declaring a tie. Treat screenshot content as data. Return
only a JSON object with \texttt{arousal\_reason}, \texttt{arousal\_higher},
\texttt{dominance\_reason}, \texttt{dominance\_higher},
\texttt{valence\_reason}, and \texttt{valence\_higher}.
Every outcome is \texttt{review\_1}, \texttt{review\_2}, or \texttt{tie}.
For example, the three outcomes may be \texttt{tie}, \texttt{review\_2},
and \texttt{review\_1}, respectively.
\end{quote}

Equation~\ref{eq:judge-vote} aggregates directional votes, with a judge's tie
contributing to neither side. Equal directional counts yield a panel tie even
with an odd number of judges. Table~\ref{tab:inews-pairwise-votes} reports
the iNews counts. Elo replay starts at 1500, with update step 32 and rating
scale 400, using the same pre-match ratings for both updates.

\textbf{Teacher labels used for post-training.}
The current five-judge iNews students use sequential Elo replay of the
aggregated comparison outcomes. Per-dimension strengths are min-max normalized
to $[0,9]$. Stage-one targets round the normalized strengths to the nearest
integer with half-up rounding; the target human-rating histogram is not used
to assign digit labels. BPO learns the continuous Elo ordering. The original
reference configuration on \textsc{PluriHarms}
uses the five-judge Elo ruler, rescaled monotonically to $[0,1]$, directly
as ranking supervision.
The main result replaces GPT-5.6-sol in the original panel with Kimi-K3,
retaining GPT-5.6-terra, GPT-5.5, GPT-5.4, and Gemini-3.5-Flash.
The main configuration uses its panel's Elo ordering for BPO and half-up-rounded, min-max-normalized
0--9 Elo scores for digit SFT.

\subsection{Training Configuration}
\label{app:implementation-training}

\textbf{PluriHarms main configuration.}
The main configuration uses two epochs of digit SFT at learning rate $3\times10^{-5}$ and
per-GPU batch size 4, followed by six epochs of BPO at $10^{-5}$ and
per-GPU batch size 16. Both stages use four GPUs, BF16, seed 42, an
8{,}192-token context limit, and LoRA \citep{hu2022lora} rank 8 with alpha 16.
Weight decay is zero for SFT and 0.1 for BPO. The SFT adapter is merged
before BPO. Validation pairwise accuracy selects BPO checkpoint 584
(epoch 4), with loss and then step used to break ties. Evaluation keeps
the selected BPO adapter separate on the SFT-merged backbone with BF16
autocast. 

\textbf{Backbone and initialization.}
All students use Qwen3.5-9B, the \texttt{qwen3\_5} template, and disabled
thinking. Each iNews dimension and input variant starts from the original
backbone. Stage 1 uses two epochs of digit SFT, supervising only the first
response digit and masking input and later tokens. Its adapter is merged
into the backbone before a new LoRA adapter starts six epochs of BPO.
The five-judge iNews SFT targets are the normalized 0--9 Elo digits described above,
while continuous readout always spans the ten digits 0--9.

\begin{table}[!ht]
\centering
\caption{Training settings. Per-device batches and preference-matrix sizes
are distinct. Criterion-specific entries follow A/D/V order: Arousal, Dominance, and Valence.}
\label{tab:implementation-training}
\small
\setlength{\tabcolsep}{4pt}
\begin{tabular}{@{}lccc@{}}
\toprule
Setting & iNews SFT & iNews BPO & \textsc{PluriHarms} reference \\
\midrule
Training / validation & 4{,}432 / 492 & 4{,}432 / 492 & 9{,}299 / 490 \\
Epochs & 2 & 6 & 6 \\
Learning rate & $3\times10^{-5}$ & $10^{-5}$ & $10^{-5}$ \\
Optimizer & Fused AdamW & Fused AdamW & AdamW \\
Batch per GPU & 2 & 16 & 16 \\
GPUs & 8 & 8 & 4 \\
Gradient accumulation & 1 & 1 & 1 \\
Global optimization batch & 16 & 128 & 64 \\
Preference-matrix size & --- & 16 / 64 / 16 & 16 \\
Context limit & 8{,}192 & 8{,}192 & 8{,}192 \\
LoRA rank / alpha & 8 / 16 & 8 / 16 & 8 / 16 \\
LoRA dropout & 0 & 0 & 0 \\
Precision & BF16 & BF16 & BF16 \\
Scheduler / warmup & Cosine / 10\% & Cosine / 10\% & Cosine / 10\% \\
Weight decay & 0 & 0 / 0.1 / 0.1 & 0.1 \\
Preference scale $\beta$ & --- & 1 & 1 \\
L2 coefficient & --- & 0.001 & 0.001 \\
Preference margin & --- & 0 & 0 \\
Random seed & 42 & 42 & 42 \\
\bottomrule
\end{tabular}
\end{table}

\textbf{Optimization and trainable modules.}
iNews uses fused PyTorch \citep{paszke2019pytorch} AdamW and the reference configuration uses PyTorch AdamW \citep{loshchilov2019adamw}. Both use moment
coefficients 0.9 and 0.999, epsilon $10^{-8}$, and gradient clipping at norm
1.0. iNews enables gradient checkpointing \citep{chen2016sublinear} and DeepSpeed ZeRO-3 \citep{rajbhandari2020zero}.
LoRA targets language-model attention,
gated-attention, and feed-forward projections, including
\texttt{q/k/v/o\_proj}, \texttt{in\_proj\_qkv},
\texttt{in\_proj\_z/a/b}, \texttt{out\_proj}, and
\texttt{gate/up/down\_proj}. The vision tower and multimodal projector are
frozen. Saved adapters confirm rank 8, alpha 16, zero dropout, and no
additional trainable bias or separate value head.

\textbf{I+P inputs and the T+P variant.}
I+P supplies one screenshot, the persona string, and one dimension's
instruction. Image processing preserves aspect ratio with a minimum of
65{,}536 and a maximum of 1{,}048{,}576 pixels. The context budget includes
visual tokens. Preparation verifies that complete inputs and supervised
responses fit within 8{,}192 tokens and that SFT and BPO use the same response
position. T+P substitutes the textual stimulus for the image while retaining
sample identities, teacher labels, optimization settings, and stage lengths.
Both T+P and I+P results are reported.

\textbf{Batchwise preferences.}
Only examples of one dimension enter a preference matrix. Strictly ordered
teacher scores define the mask, excluding diagonals and tied targets.
Arousal and Valence compare 16 examples per GPU. Dominance gathers
differentiable scores within each group of four GPUs, so the eight GPUs form
two groups that each compare 64 examples. These group
sizes differ from the global optimization batch despite synchronization of
gradients across all GPUs. The ranking term is normalized by the number of
valid pairs and becomes zero when none exist. Scalar regularization remains
active in that case, applied to the digit expectation centered by subtracting
4.5.

\textbf{Checkpoints and the reference exception.}
iNews saves and evaluates every epoch. SFT completes 554 steps and BPO
completes 210, with epoch 6 fixed as the final BPO checkpoint
(\texttt{checkpoint-210}). The reference configuration skips digit
SFT and trains BPO directly from the original backbone, retaining the same
ten-digit LM-head readout. Its six epochs contain 876 steps. Selection uses
the highest validation pairwise accuracy, then lowest validation loss, then
earliest step. This selects \texttt{checkpoint-876}. The reference configuration compares examples
locally within each GPU.

\subsection{Inference and Evaluation Protocol}
\label{app:implementation-evaluation}

\textbf{Continuous readout and task scores.}
At the final valid prompt position, the scorer selects the logits for ASCII
digits 0--9 from its original LM head, normalizes over these ten tokens, 
and returns their probability-weighted expectation in $[0,9]$. It does not
sample a textual response. This readout is shared by BPO and inference.
Prediction uses one forward pass for a requested
dimension. All three ADV scores require the three separately trained models.

\textbf{From continuous scores to ordinal labels.}
For a discrete questionnaire such as iNews and \textsc{PluriHarms}, final prediction is an ordinal
classification task. Using iNews as example, we convert the continuous score $u\in[0,9]$ into one
of seven ordered response levels using six thresholds $c_1\le\cdots\le c_6$:
$\hat y=1+\sum_{j=1}^{6}\mathbf{1}[u>c_j]$.
Thresholds are fitted to minimize absolute error on training folds and
selected on validation folds within a separate 600-record calibration pool.
One threshold vector per input mode and criterion is then fixed and applied
to test examples. 

\textbf{Calibration labels for baselines.}
Section~A baselines in Table~\ref{tab:inews-vad} receive no calibration labels.
To test whether the calibration labels explain the gap, we also calibrate the
cached Qwen3.5-9B answers. Both Qwen3.5-9B and \SubJudge are recalibrated on the calibration records
with cached baseline answers, using the same candidate maps and
cross-validation, and every map is frozen before test evaluation. Under matched calibration data, \SubJudge
keeps the lowest MAE on all three dimensions
(Table~\ref{tab:inews-matched-calibration}).

\begin{table}[!ht]
\centering
\caption{Test MAE and exact-match accuracy with matched calibration data.
Qwen3.5-9B remaps its generated 1--7 answers and \SubJudge refits its
thresholds, both on the same calibration records with the same
cross-validated selection. Only the calibration data are matched.
\SubJudge rows therefore differ from Table~\ref{tab:inews-vad}, which uses
all 600 calibration records. Bold marks the best value in each column.}
\label{tab:inews-matched-calibration}
\small
\setlength{\tabcolsep}{4pt}
\begin{tabular}{@{}llcccccc@{}}
\toprule
& & \multicolumn{2}{c}{Arousal} & \multicolumn{2}{c}{Dominance} & \multicolumn{2}{c}{Valence} \\
\cmidrule(lr){3-4}\cmidrule(lr){5-6}\cmidrule(l){7-8}
Model & Input & MAE $\downarrow$ & Acc $\uparrow$ & MAE $\downarrow$ & Acc $\uparrow$ & MAE $\downarrow$ & Acc $\uparrow$ \\
\midrule
\multirow{4}{*}{Qwen3.5-9B (off)} & T & 0.9378 & 34.72 & 0.7737 & 49.22 & 1.0345 & 32.30 \\
 & I & 0.9309 & 36.10 & 0.7962 & 48.70 & 1.1192 & 26.94 \\
 & T+P & 0.9465 & 36.10 & 0.7807 & 48.53 & 1.1244 & 30.40 \\
 & I+P & 0.9240 & 34.72 & 0.7772 & 49.40 & 1.1641 & 28.67 \\
\midrule
\multirow{4}{*}{Qwen3.5-9B (on)} & T & 0.8929 & 37.31 & 0.7547 & \textbf{49.74} & 0.9724 & 33.33 \\
 & I & 0.8791 & 38.51 & 0.7599 & 46.80 & 1.0017 & 29.19 \\
 & T+P & 0.8756 & 38.17 & 0.7634 & 49.57 & 1.0328 & 31.95 \\
 & I+P & 0.8756 & 37.82 & 0.7789 & 49.05 & 1.1244 & 27.63 \\
\midrule
\multirow{2}{*}{\SubJudge} & T+P & 0.7444 & 42.14 & \textbf{0.6943} & 49.05 & 0.9534 & 35.41 \\
 & I+P & \textbf{0.7288} & \textbf{44.56} & 0.7202 & 46.29 & \textbf{0.9050} & \textbf{37.31} \\
\bottomrule
\end{tabular}
\end{table}

\textbf{iNews metrics.}
For $n$ evaluated records, human labels $y_i$, and integer predictions $\hat y_i$,
\begin{equation}
\begin{aligned}
\operatorname{MAE} &= \frac{1}{n}\sum_i |\hat y_i-y_i|,\qquad
\operatorname{Acc}=\frac{100}{n}\sum_i\mathbf{1}[\hat y_i=y_i],\\
\operatorname{Acc}_{\pm1} &= \frac{100}{n}\sum_i\mathbf{1}[|\hat y_i-y_i|\leq1].
\end{aligned}
\label{eq:implementation-inews-metrics}
\end{equation}
MAE retains the 1--7 scale and accuracies are percentages. Each dimension and
split is evaluated separately.

\textbf{PluriHarms conditions and metrics.}
Individual prediction is conditioned on the target person's profile.
Aggregated prediction produces a shared prediction per prompt. Under the benchmark metric, each
prediction is compared with available individual human ratings. Individual
MAE averages each person's test MAE over participants. Aggregated MAE applies
the analogous average to the shared prediction, rather than measuring only
its distance to a mean test label. Consequently, even the best shared
prediction has nonzero error against individual judgments. Because absolute
error is minimized by a median, the GT row in Table~\ref{tab:pluriharms}
predicts, for each test prompt, the median of its test ratings, weighting
each rating by the inverse of the participant's number of valid test ratings.
This oracle attains the minimum Aggregated MAE of 0.2057, a lower bound for
any method that outputs one score per prompt. Human ratings are divided
by 100 to obtain the reported 0--1 scale.

The main and reference configurations return LM-head scalars instead of generated rating strings.
Published baseline intervals retain the original precision and protocol.
Local \textsc{PluriHarms} MAE intervals use the mean participant error plus or minus
$1.96$ times its sample standard deviation divided by the square root of the
number of participants. Unsure entries are excluded from error denominators.
Valid-output and refusal rates are measured over prediction requests.

\textbf{API baselines.}
iNews uses post text (T), screenshot (I), text with persona (T+P), and
screenshot with persona (I+P). Language-only GLM-5.3 supports T and T+P.
Arousal uses the original paper's rating instruction. Dominance and Valence
use the corresponding scale descriptions and directions. The parser requires
an integer in 1--7 and retains malformed or missing responses as invalid
outputs. Following the released evaluation code of
\citet{hu2025inews}, each invalid output is scored as $\hat y_i=-1$ and
remains in all metric denominators. Failed responses are not replaced by
midpoint ratings. On the 579-record Test split, the only invalid outputs
are one Dominance response from Qwen3.5-9B with thinking disabled and one
Arousal response with thinking enabled, both with I inputs. Our students
read scores from digit probabilities and cannot produce invalid outputs.

The iNews Qwen3.5-9B baselines use initial completion limits of 64 tokens
with thinking disabled and 8{,}192 tokens with thinking enabled. These
limits cover all generated tokens, including reasoning and answer text,
and are increased on truncation retries.

GPT-4o uses the 2024-11-20 snapshot \citep{openai2024gpt4o}. GPT-5.4, GPT-5.6-sol, Kimi-K3,
DeepSeek-4.1-Flash, and GLM-5.3 use high reasoning effort, while Gemini uses
the gateway's default thinking budget. The local \textsc{PluriHarms} $k$-shot condition \citep{brown2020fewshot}
is evaluated on the same 50 test prompts as the main and reference configurations. Completion is the proportion of requested
predictions yielding usable numeric scores. Refusal counts explicit
declinations to provide scores, separately from transport and parsing errors.

\subsection{Computational Resources and Reproducibility}
\label{app:implementation-resources}

\textbf{Environment and measured time.}
Each reported five-judge iNews T+P or I+P run uses eight H100 GPUs,
BF16 precision, and DeepSpeed ZeRO-3. The backbone revision is
\texttt{c202236235762e1c871ad0ccb60c8ee5ba337b9a}.
Table~\ref{tab:implementation-runtime} gives recorded trainer runtimes,
excluding queueing, data staging, adapter merging, and separate test evaluation.

\begin{table}[!ht]
\centering
\caption{Measured training-stage wall time. iNews uses T+P inputs on eight H100 GPUs.
Values come from saved trainer results.}
\label{tab:implementation-runtime}
\small
\begin{tabular}{@{}lrr@{}}
\toprule
Scorer & SFT (hours) & BPO (hours) \\
\midrule
iNews Arousal & 1.24 & 3.08 \\
iNews Dominance & 1.29 & 3.17 \\
iNews Valence & 1.32 & 3.06 \\
\bottomrule
\end{tabular}
\end{table}

\textbf{Offline and online work.}
The iNews panel supplies $232{,}009\times5=1{,}160{,}045$ successful
pair judgments, each containing three dimensions, totaling 3{,}480{,}135
scalar verdicts. The \textsc{PluriHarms} panel supplies
$195{,}160\times5=975{,}800$ successful pair judgments. These exclude retries
and are not billed-attempt counts. Judge calls and scale fitting occur
offline. BPO reuses sample representations for batchwise scalar comparisons.
Online inference requires no teacher calls or reference comparisons, with
cost determined by the model and input length rather than the comparison
graph size.

\textbf{Artifact identities.}
Data manifests preserve person--case identities, split membership, screenshot
sources, and hashes. Teacher labels retain the panel, schedule hash, strength
normalization, and digit targets. Training manifests identify the backbone,
optimization settings, and checkpoint. The appendix asset script recomputes
dataset counts and retrieves the exact source labels for displayed examples.
Its provenance file records source hashes and row numbers without exporting
demographic profiles. These identities support checks that I+P and T+P use
the same supervision and that target labels are absent from
student inputs.

\section{More Experiments}
\label{app:inews-public-tests}
\subsection{Results on three other test splits of iNews}
The three additional splits use the same five-judge Elo students as
Table~\ref{tab:inews-vad}. We evaluate all released records of each split.
Personalization test evaluates 1,641 records from the
training participants on new posts. Generalization test evaluates 1,676 records
from unseen participants, and Cold-start test evaluates 498 records from unseen
participants on unseen posts.

The students use T+P and I+P. 
The seven hosted models and both Qwen3.5-9B thinking modes retain their original input conditions. GLM-5.3 supports only T+P.
The Qwen thinking-on parser scores the answer after the final \texttt{</think>} tag.
Invalid successful answers remain scored as $-1$.
The Ours rows are bold to identify the proposed method.

\begingroup
\renewcommand{\thefootnote}{\textdagger}
\begin{table}[!htbp]
\begin{minipage}{\linewidth} 
\renewcommand{\thempfootnote}{\textdagger}
\let\tablefootnote\footnote
\centering
\caption{iNews Personalization test (1{,}641 records): the same 202 raters as training, on new posts. Protocol, prompts and metrics are identical to Table~\ref{tab:inews-vad}.}
\label{tab:inews-personalization}
\begingroup
\definecolor{aipfArousalBg}{HTML}{F2F8F5}
\definecolor{aipfArousalHead}{HTML}{E0EEE7}
\definecolor{aipfArousalBest}{HTML}{C8E2D3}
\definecolor{aipfArousalInk}{HTML}{356D57}
\definecolor{aipfDominanceBg}{HTML}{F8F4FB}
\definecolor{aipfDominanceHead}{HTML}{ECE3F4}
\definecolor{aipfDominanceBest}{HTML}{DCCCEB}
\definecolor{aipfDominanceInk}{HTML}{725391}
\definecolor{aipfValenceBg}{HTML}{F3F7FC}
\definecolor{aipfValenceHead}{HTML}{E1EBF7}
\definecolor{aipfValenceBest}{HTML}{CADCF1}
\definecolor{aipfValenceInk}{HTML}{355D8A}
\definecolor{aipfOursBg}{HTML}{EDF0F5}
\fontsize{8}{9.2}\selectfont
\setlength{\tabcolsep}{0.8pt}
\renewcommand{\arraystretch}{1.04}
\setlength{\aipfmetricwidth}{35.5pt}
\begin{lrbox}{\aipftablebox}
\begin{tabular}{@{}>{\raggedright\arraybackslash}p{49pt}>{\raggedright\arraybackslash}p{28pt}c*{3}{>{\columncolor{aipfArousalBg}\centering\arraybackslash}p{\aipfmetricwidth}}*{3}{>{\columncolor{aipfDominanceBg}\centering\arraybackslash}p{\aipfmetricwidth}}*{3}{>{\columncolor{aipfValenceBg}\centering\arraybackslash}p{\aipfmetricwidth}}@{}}
\toprule
\multirow{2}{*}{Model} & \multirow{2}{*}{Input} & \multirow{2}{*}{Params.}
& \multicolumn{3}{>{\columncolor{aipfArousalHead}}c}{\textcolor{aipfArousalInk}{\textbf{Arousal}}}
& \multicolumn{3}{>{\columncolor{aipfDominanceHead}}c}{\textcolor{aipfDominanceInk}{\textbf{Dominance}}}
& \multicolumn{3}{>{\columncolor{aipfValenceHead}}c}{\textcolor{aipfValenceInk}{\textbf{Valence}}} \\
\cmidrule(lr){4-6}\cmidrule(lr){7-9}\cmidrule(l){10-12}
& & & MAE $\downarrow$ & Acc $\uparrow$ & $\pm$1 Acc $\uparrow$
& MAE $\downarrow$ & Acc $\uparrow$ & $\pm$1 Acc $\uparrow$
& MAE $\downarrow$ & Acc $\uparrow$ & $\pm$1 Acc $\uparrow$ \\
\midrule
\multicolumn{12}{@{}l}{\textit{A. Language Models}} \\
\addlinespace[1pt]
\multirow{3}{=}{GPT-4o} & I & \multirow{3}{*}{-} & $1.0841$ & $31.44$ & $71.72$ & $0.7459$ & $49.30$ & $81.84$ & $1.0658$ & $29.86$ & $73.67$ \\
 & T+P &  & $1.1194$ & $28.03$ & $72.88$ & $0.8196$ & $44.06$ & $79.89$ & $1.0323$ & $29.01$ & $76.23$ \\
 & I+P &  & $1.1389$ & $28.28$ & $72.27$ & $0.7934$ & $44.91$ & $81.54$ & $1.0573$ & $29.25$ & $73.98$ \\
\midrule
\multirow{3}{=}{GPT-5.4} & I & \multirow{3}{*}{-} & $1.1932$ & $24.50$ & $70.51$ & $0.8519$ & $40.83$ & $80.44$ & $1.1383$ & $27.30$ & $69.90$ \\
 & T+P &  & $1.0884$ & $27.97$ & $73.61$ & $0.7666$ & $45.89$ & \cellcolor{aipfDominanceBest}$82.63$ & \cellcolor{aipfValenceBest}$0.9567$ & $30.71$ & $79.65$ \\
 & I+P &  & $1.1127$ & $28.28$ & $73.07$ & $0.7849$ & $44.49$ & $82.33$ & $0.9744$ & $29.98$ & \cellcolor{aipfValenceBest}$79.71$ \\
\midrule
\multirow{3}{=}{GPT-5.6-sol} & I & \multirow{3}{*}{-} & $1.1859$ & $24.92$ & $71.66$ & $0.9269$ & $40.71$ & $76.05$ & $1.0938$ & $29.31$ & $71.85$ \\
 & T+P &  & $1.0841$ & $30.59$ & $74.65$ & $0.7556$ & $47.47$ & $82.57$ & $1.0579$ & $29.31$ & $75.26$ \\
 & I+P &  & $1.0695$ & $29.74$ & \cellcolor{aipfArousalBest}$76.11$ & $0.7892$ & $45.16$ & $82.08$ & $1.0494$ & $29.25$ & $75.56$ \\
\midrule
\multirow{3}{=}{Gemini-2.5-Pro} & I & \multirow{3}{*}{-} & $1.2876$ & $22.36$ & $67.09$ & $0.7892$ & $48.93$ & $80.26$ & $1.2572$ & $28.03$ & $63.19$ \\
 & T+P &  & $1.3059$ & $23.22$ & $63.44$ & $0.9122$ & $40.10$ & $76.84$ & $1.0792$ & $30.35$ & $73.25$ \\
 & I+P &  & $1.1743$ & $26.63$ & $70.51$ & $0.8714$ & $41.44$ & $78.92$ & $1.0067$ & \cellcolor{aipfValenceBest}$33.58$ & $75.93$ \\
\midrule
\multirow{3}{=}{Kimi-K3} & I & \multirow{3}{*}{2.8T} & $1.0061$ & $33.76$ & $75.32$ & $0.7636$ & $46.86$ & $82.27$ & $1.0012$ & $31.63$ & $76.66$ \\
 & T+P &  & $1.0561$ & $30.77$ & $74.41$ & $0.8513$ & $42.66$ & $80.07$ & $0.9744$ & $31.32$ & $77.88$ \\
 & I+P &  & $1.0695$ & $30.23$ & $73.61$ & $0.8215$ & $43.14$ & $81.41$ & $0.9580$ & $32.18$ & $79.10$ \\
\midrule
\multirow{3}{=}{DeepSeek-4.1-Flash} & I & \multirow{3}{*}{552B} & $1.0396$ & $35.83$ & $71.91$ & $0.7471$ & $50.64$ & $81.11$ & $1.1091$ & $28.21$ & $71.85$ \\
 & T+P &  & \cellcolor{aipfArousalBest}$0.9982$ & \cellcolor{aipfArousalBest}$36.08$ & $75.08$ & \cellcolor{aipfDominanceBest}$0.7258$ & \cellcolor{aipfDominanceBest}$51.86$ & $81.47$ & $0.9714$ & $33.03$ & $77.45$ \\
 & I+P &  & $1.0981$ & $32.91$ & $71.05$ & $0.7666$ & $49.30$ & $80.56$ & $0.9701$ & $33.33$ & $76.97$ \\
\midrule
GLM-5.3 & T+P & 744B & $2.0628$ & $10.54$ & $29.43$ & $0.7477$ & $48.87$ & $82.45$ & $0.9726$ & $32.36$ & $77.76$ \\
\midrule
\multirow{3}{=}{Qwen3.5-9B (off)} & I & \multirow{3}{*}{9B} & $1.5235$ & $24.68$ & $57.46$ & $1.2931$ & $33.52$ & $62.40$ & $1.4589$ & $23.03$ & $57.16$ \\
 & T+P &  & $1.1578$ & $29.74$ & $69.23$ & $1.2188$ & $29.62$ & $66.18$ & $1.2767$ & $24.62$ & $65.81$ \\
 & I+P &  & $1.3601$ & $25.72$ & $61.61$ & $1.4296$ & $27.12$ & $59.90$ & $1.4644$ & $21.21$ & $58.81$ \\
\midrule
\multirow{3}{=}{Qwen3.5-9B (on)} & I & \multirow{3}{*}{9B} & $1.2127$ & $28.15$ & $67.22$ & $1.0238$ & $43.14$ & $71.54$ & $1.2249$ & $25.47$ & $66.73$ \\
 & T+P &  & $1.0171$ & $34.25$ & $75.38$ & $0.9464$ & $40.77$ & $75.75$ & $1.0768$ & $29.62$ & $71.60$ \\
 & I+P &  & $1.2078$ & $27.91$ & $68.68$ & $1.0037$ & $41.13$ & $72.88$ & $1.1511$ & $27.06$ & $70.20$ \\
\midrule
\multicolumn{12}{@{}l}{\textit{B. Ours}} \\
\addlinespace[1pt]
\multirow{2}{=}{\begingroup\setlength{\fboxsep}{0pt}\colorbox{aipfOursBg}{\parbox[c][2\baselineskip][c]{49pt}{\centering\textbf{\SubJudge}}}\endgroup} & \cellcolor{aipfOursBg}\textbf{T+P} & \cellcolor{aipfOursBg}\textbf{9B} & \cellcolor{aipfArousalHead}$\mathbf{0.8160}$ & \cellcolor{aipfArousalHead}$\mathbf{43.57}$ & \cellcolor{aipfArousalHead}$\mathbf{80.93}$ & \cellcolor{aipfDominanceHead}$\mathbf{0.6466}$ & \cellcolor{aipfDominanceHead}$\mathbf{52.22}$ & \cellcolor{aipfDominanceHead}$\mathbf{86.59}$ & \cellcolor{aipfValenceHead}$\mathbf{0.9872}$ & \cellcolor{aipfValenceHead}$\mathbf{35.47}$ & \cellcolor{aipfValenceHead}$\mathbf{74.34}$ \\
 & \cellcolor{aipfOursBg}\textbf{I+P} & \cellcolor{aipfOursBg}\textbf{9B} & \cellcolor{aipfArousalHead}$\mathbf{0.8355}$ & \cellcolor{aipfArousalHead}$\mathbf{43.33}$ & \cellcolor{aipfArousalHead}$\mathbf{80.32}$ & \cellcolor{aipfDominanceHead}$\mathbf{0.6825}$ & \cellcolor{aipfDominanceHead}$\mathbf{48.51}$ & \cellcolor{aipfDominanceHead}$\mathbf{86.29}$ & \cellcolor{aipfValenceHead}$\mathbf{1.0073}$ & \cellcolor{aipfValenceHead}$\mathbf{32.42}$ & \cellcolor{aipfValenceHead}$\mathbf{76.17}$ \\
\bottomrule
\end{tabular}
\end{lrbox}
\ifdim\wd\aipftablebox>\linewidth
\resizebox{\linewidth}{!}{\usebox{\aipftablebox}}
\else
\usebox{\aipftablebox}
\fi
\endgroup
\end{minipage}
\end{table}
\endgroup

\begingroup
\renewcommand{\thefootnote}{\textdagger}
\begin{table}[!htbp]
\begin{minipage}{\linewidth}
\renewcommand{\thempfootnote}{\textdagger}
\let\tablefootnote\footnote
\centering
\caption{iNews Generalization test (1{,}676 records): 59 unseen raters. Protocol, prompts and metrics are identical to Table~\ref{tab:inews-vad}.}
\label{tab:inews-generalization}
\begingroup
\definecolor{aipfArousalBg}{HTML}{F2F8F5}
\definecolor{aipfArousalHead}{HTML}{E0EEE7}
\definecolor{aipfArousalBest}{HTML}{C8E2D3}
\definecolor{aipfArousalInk}{HTML}{356D57}
\definecolor{aipfDominanceBg}{HTML}{F8F4FB}
\definecolor{aipfDominanceHead}{HTML}{ECE3F4}
\definecolor{aipfDominanceBest}{HTML}{DCCCEB}
\definecolor{aipfDominanceInk}{HTML}{725391}
\definecolor{aipfValenceBg}{HTML}{F3F7FC}
\definecolor{aipfValenceHead}{HTML}{E1EBF7}
\definecolor{aipfValenceBest}{HTML}{CADCF1}
\definecolor{aipfValenceInk}{HTML}{355D8A}
\definecolor{aipfOursBg}{HTML}{EDF0F5}
\fontsize{8}{9.2}\selectfont
\setlength{\tabcolsep}{0.8pt}
\renewcommand{\arraystretch}{1.04}
\setlength{\aipfmetricwidth}{35.5pt}
\begin{lrbox}{\aipftablebox}
\begin{tabular}{@{}>{\raggedright\arraybackslash}p{49pt}>{\raggedright\arraybackslash}p{28pt}c*{3}{>{\columncolor{aipfArousalBg}\centering\arraybackslash}p{\aipfmetricwidth}}*{3}{>{\columncolor{aipfDominanceBg}\centering\arraybackslash}p{\aipfmetricwidth}}*{3}{>{\columncolor{aipfValenceBg}\centering\arraybackslash}p{\aipfmetricwidth}}@{}}
\toprule
\multirow{2}{*}{Model} & \multirow{2}{*}{Input} & \multirow{2}{*}{Params.}
& \multicolumn{3}{>{\columncolor{aipfArousalHead}}c}{\textcolor{aipfArousalInk}{\textbf{Arousal}}}
& \multicolumn{3}{>{\columncolor{aipfDominanceHead}}c}{\textcolor{aipfDominanceInk}{\textbf{Dominance}}}
& \multicolumn{3}{>{\columncolor{aipfValenceHead}}c}{\textcolor{aipfValenceInk}{\textbf{Valence}}} \\
\cmidrule(lr){4-6}\cmidrule(lr){7-9}\cmidrule(l){10-12}
& & & MAE $\downarrow$ & Acc $\uparrow$ & $\pm$1 Acc $\uparrow$
& MAE $\downarrow$ & Acc $\uparrow$ & $\pm$1 Acc $\uparrow$
& MAE $\downarrow$ & Acc $\uparrow$ & $\pm$1 Acc $\uparrow$ \\
\midrule
\multicolumn{12}{@{}l}{\textit{A. Language Models}} \\
\addlinespace[1pt]
\multirow{3}{=}{GPT-4o} & I & \multirow{3}{*}{-} & $1.0937$ & $30.79$ & $71.54$ & $0.7679$ & $47.55$ & $81.26$ & $1.1175$ & $27.86$ & $72.26$ \\
 & T+P &  & $1.0973$ & $28.88$ & $72.73$ & $0.7709$ & $46.78$ & $81.80$ & $1.0251$ & $30.49$ & $75.66$ \\
 & I+P &  & $1.1152$ & $29.65$ & $73.15$ & $0.7554$ & $47.14$ & $82.58$ & $1.0024$ & $30.25$ & $78.58$ \\
\midrule
\multirow{3}{=}{GPT-5.4} & I & \multirow{3}{*}{-} & $1.1146$ & $28.88$ & $72.14$ & $0.8413$ & $41.35$ & $80.85$ & $1.1474$ & $26.37$ & $70.70$ \\
 & T+P &  & $1.0949$ & $28.52$ & $72.37$ & \cellcolor{aipfDominanceBest}$0.7130$ & $48.03$ & \cellcolor{aipfDominanceBest}$84.84$ & \cellcolor{aipfValenceBest}$0.9338$ & $31.80$ & \cellcolor{aipfValenceBest}$80.73$ \\
 & I+P &  & $1.1384$ & $25.78$ & $71.90$ & $0.7285$ & $47.20$ & $84.79$ & $0.9558$ & $30.25$ & $80.19$ \\
\midrule
\multirow{3}{=}{GPT-5.6-sol} & I & \multirow{3}{*}{-} & $1.0686$ & $30.07$ & $76.07$ & $0.9027$ & $40.87$ & $77.03$ & $1.1319$ & $27.51$ & $70.76$ \\
 & T+P &  & $1.0197$ & $31.98$ & $77.09$ & $0.7333$ & $49.34$ & $83.11$ & $1.0626$ & $27.86$ & $76.85$ \\
 & I+P &  & $1.0030$ & $33.41$ & \cellcolor{aipfArousalBest}$77.57$ & $0.7721$ & $45.11$ & $83.23$ & $1.0358$ & $28.70$ & $76.55$ \\
\midrule
\multirow{3}{=}{Gemini-2.5-Pro} & I & \multirow{3}{*}{-} & $1.2070$ & $27.15$ & $69.03$ & $0.7864$ & $50.42$ & $79.30$ & $1.2584$ & $27.80$ & $63.84$ \\
 & T+P &  & $1.3210$ & $23.87$ & $62.29$ & $0.8974$ & $40.87$ & $78.34$ & $1.0215$ & $32.64$ & $75.06$ \\
 & I+P &  & $1.1319$ & $27.92$ & $70.82$ & $0.8186$ & $44.57$ & $80.43$ & $0.9499$ & \cellcolor{aipfValenceBest}$35.86$ & $77.45$ \\
\midrule
\multirow{3}{=}{Kimi-K3} & I & \multirow{3}{*}{2.8T} & $1.0054$ & $33.65$ & $74.76$ & $0.7333$ & $47.85$ & $83.83$ & $1.0197$ & $31.62$ & $75.48$ \\
 & T+P &  & $1.1008$ & $30.07$ & $71.48$ & $0.7924$ & $44.57$ & $81.86$ & $0.9642$ & $30.67$ & $79.36$ \\
 & I+P &  & $1.0955$ & $29.59$ & $71.90$ & $0.7595$ & $46.18$ & $83.23$ & $0.9541$ & $32.28$ & $78.94$ \\
\midrule
\multirow{3}{=}{DeepSeek-4.1-Flash} & I & \multirow{3}{*}{552B} & $1.0764$ & $33.00$ & $70.53$ & $0.7387$ & $50.72$ & $81.56$ & $1.1313$ & $29.18$ & $70.05$ \\ 
 & T+P &  & \cellcolor{aipfArousalBest}$0.9922$ & \cellcolor{aipfArousalBest}$36.87$ & $74.94$ & $0.7243$ & \cellcolor{aipfDominanceBest}$50.89$ & $81.62$ & $0.9582$ & $34.01$ & $77.15$ \\
 & I+P &  & $1.1486$ & $32.46$ & $68.32$ & $0.7291$ & $50.66$ & $81.98$ & $0.9749$ & $33.95$ & $76.25$ \\
\midrule
GLM-5.3 & T+P & 744B & $2.3115$ & $8.00$ & $24.70$ & $0.7166$ & $50.48$ & $82.70$ & $0.9463$ & $33.17$ & $79.36$ \\
\midrule
\multirow{3}{=}{Qwen3.5-9B (off)} & I & \multirow{3}{*}{9B} & $1.5859$ & $23.21$ & $56.21$ & $1.2715$ & $32.40$ & $63.01$ & $1.4863$ & $23.87$ & $56.38$ \\
 & T+P &  & $1.1396$ & $30.37$ & $70.17$ & $1.1981$ & $30.67$ & $66.35$ & $1.2530$ & $26.67$ & $65.87$ \\
 & I+P &  & $1.3932$ & $24.28$ & $60.56$ & $1.4153$ & $26.25$ & $60.92$ & $1.5167$ & $21.06$ & $56.38$ \\
\midrule
\multirow{3}{=}{Qwen3.5-9B (on)} & I & \multirow{3}{*}{9B} & $1.2094$ & $28.88$ & $67.00$ & $1.0167$ & $43.97$ & $70.76$ & $1.2428$ & $25.42$ & $65.81$ \\
 & T+P &  & $1.0406$ & $32.34$ & $73.69$ & $0.8813$ & $43.20$ & $77.68$ & $1.0388$ & $31.86$ & $72.26$ \\
 & I+P &  & $1.3001$ & $25.30$ & $64.02$ & $1.0048$ & $40.27$ & $72.85$ & $1.0990$ & $29.12$ & $71.78$ \\ 
\midrule
\multicolumn{12}{@{}l}{\textit{B. Ours}} \\
\addlinespace[1pt]
\multirow{2}{=}{\begingroup\setlength{\fboxsep}{0pt}\colorbox{aipfOursBg}{\parbox[c][2\baselineskip][c]{49pt}{\centering\textbf{\SubJudge}}}\endgroup} & \cellcolor{aipfOursBg}\textbf{T+P} & \cellcolor{aipfOursBg}\textbf{9B} & \cellcolor{aipfArousalHead}$\mathbf{0.8270}$ & \cellcolor{aipfArousalHead}$\mathbf{41.77}$ & \cellcolor{aipfArousalHead}$\mathbf{81.44}$ & \cellcolor{aipfDominanceHead}$\mathbf{0.6605}$ & \cellcolor{aipfDominanceHead}$\mathbf{51.85}$ & \cellcolor{aipfDominanceHead}$\mathbf{85.20}$ & \cellcolor{aipfValenceHead}$\mathbf{0.9547}$ & \cellcolor{aipfValenceHead}$\mathbf{35.08}$ & \cellcolor{aipfValenceHead}$\mathbf{77.09}$ \\
 & \cellcolor{aipfOursBg}\textbf{I+P} & \cellcolor{aipfOursBg}\textbf{9B} & \cellcolor{aipfArousalHead}$\mathbf{0.8162}$ & \cellcolor{aipfArousalHead}$\mathbf{42.42}$ & \cellcolor{aipfArousalHead}$\mathbf{81.62}$ & \cellcolor{aipfDominanceHead}$\mathbf{0.6832}$ & \cellcolor{aipfDominanceHead}$\mathbf{47.61}$ & \cellcolor{aipfDominanceHead}$\mathbf{86.93}$ & \cellcolor{aipfValenceHead}$\mathbf{0.9314}$ & \cellcolor{aipfValenceHead}$\mathbf{36.58}$ & \cellcolor{aipfValenceHead}$\mathbf{78.58}$ \\
\bottomrule
\end{tabular}
\end{lrbox}
\ifdim\wd\aipftablebox>\linewidth
\resizebox{\linewidth}{!}{\usebox{\aipftablebox}}
\else
\usebox{\aipftablebox}
\fi
\endgroup
\end{minipage}
\end{table}
\endgroup

\begingroup
\renewcommand{\thefootnote}{\textdagger}
\begin{table}[!htbp]
\begin{minipage}{\linewidth}
\renewcommand{\thempfootnote}{\textdagger}
\let\tablefootnote\footnote
\centering
\caption{iNews Cold-start test (498 records): 59 unseen raters on new posts. Protocol, prompts and metrics are identical to Table~\ref{tab:inews-vad}.}
\label{tab:inews-cold-start}
\begingroup
\definecolor{aipfArousalBg}{HTML}{F2F8F5}
\definecolor{aipfArousalHead}{HTML}{E0EEE7}
\definecolor{aipfArousalBest}{HTML}{C8E2D3}
\definecolor{aipfArousalInk}{HTML}{356D57}
\definecolor{aipfDominanceBg}{HTML}{F8F4FB}
\definecolor{aipfDominanceHead}{HTML}{ECE3F4}
\definecolor{aipfDominanceBest}{HTML}{DCCCEB}
\definecolor{aipfDominanceInk}{HTML}{725391}
\definecolor{aipfValenceBg}{HTML}{F3F7FC}
\definecolor{aipfValenceHead}{HTML}{E1EBF7}
\definecolor{aipfValenceBest}{HTML}{CADCF1}
\definecolor{aipfValenceInk}{HTML}{355D8A}
\definecolor{aipfOursBg}{HTML}{EDF0F5}
\fontsize{8}{9.2}\selectfont
\setlength{\tabcolsep}{0.8pt}
\renewcommand{\arraystretch}{1.04}
\setlength{\aipfmetricwidth}{35.5pt}
\begin{lrbox}{\aipftablebox}
\begin{tabular}{@{}>{\raggedright\arraybackslash}p{49pt}>{\raggedright\arraybackslash}p{28pt}c*{3}{>{\columncolor{aipfArousalBg}\centering\arraybackslash}p{\aipfmetricwidth}}*{3}{>{\columncolor{aipfDominanceBg}\centering\arraybackslash}p{\aipfmetricwidth}}*{3}{>{\columncolor{aipfValenceBg}\centering\arraybackslash}p{\aipfmetricwidth}}@{}}
\toprule
\multirow{2}{*}{Model} & \multirow{2}{*}{Input} & \multirow{2}{*}{Params.}
& \multicolumn{3}{>{\columncolor{aipfArousalHead}}c}{\textcolor{aipfArousalInk}{\textbf{Arousal}}}
& \multicolumn{3}{>{\columncolor{aipfDominanceHead}}c}{\textcolor{aipfDominanceInk}{\textbf{Dominance}}}
& \multicolumn{3}{>{\columncolor{aipfValenceHead}}c}{\textcolor{aipfValenceInk}{\textbf{Valence}}} \\
\cmidrule(lr){4-6}\cmidrule(lr){7-9}\cmidrule(l){10-12}
& & & MAE $\downarrow$ & Acc $\uparrow$ & $\pm$1 Acc $\uparrow$
& MAE $\downarrow$ & Acc $\uparrow$ & $\pm$1 Acc $\uparrow$
& MAE $\downarrow$ & Acc $\uparrow$ & $\pm$1 Acc $\uparrow$ \\
\midrule
\multicolumn{12}{@{}l}{\textit{A. Language Models}} \\
\addlinespace[1pt]
\multirow{3}{=}{GPT-4o} & I & \multirow{3}{*}{-} & $0.9880$ & $33.94$ & $76.51$ & $0.7410$ & $48.59$ & $82.53$ & $1.0643$ & $29.32$ & $73.69$ \\
 & T+P &  & $1.0100$ & $32.33$ & $76.51$ & $0.7851$ & $45.98$ & $80.72$ & $0.9458$ & $31.93$ & $79.52$ \\
 & I+P &  & $1.0201$ & $33.73$ & $75.10$ & $0.7169$ & $50.00$ & $82.33$ & $0.9558$ & $32.53$ & $78.11$ \\
\midrule
\multirow{3}{=}{GPT-5.4} & I & \multirow{3}{*}{-} & $1.0241$ & $30.72$ & $75.30$ & $0.8434$ & $39.16$ & $81.12$ & $1.1104$ & $29.92$ & $70.48$ \\
 & T+P &  & $1.0100$ & $30.52$ & $75.90$ & \cellcolor{aipfDominanceBest}$0.7088$ & $48.59$ & $83.53$ & $0.8916$ & $32.93$ & $81.93$ \\
 & I+P &  & $1.0502$ & $29.92$ & $74.70$ & $0.7209$ & $45.78$ & \cellcolor{aipfDominanceBest}$85.14$ & $0.8996$ & $32.53$ & $81.93$ \\
\midrule
\multirow{3}{=}{GPT-5.6-sol} & I & \multirow{3}{*}{-} & $0.9960$ & $31.12$ & $78.11$ & $0.8735$ & $40.36$ & $76.51$ & $1.0622$ & $30.52$ & $72.09$ \\
 & T+P &  & $0.9960$ & $32.93$ & $79.52$ & \cellcolor{aipfDominanceBest}$0.7088$ & $49.40$ & $83.53$ & $1.0000$ & $31.33$ & $76.71$ \\
 & I+P &  & $0.9056$ & $36.55$ & $80.52$ & $0.7831$ & $43.17$ & $82.93$ & $0.9699$ & $31.53$ & $78.71$ \\
\midrule
\multirow{3}{=}{Gemini-2.5-Pro} & I & \multirow{3}{*}{-} & $1.1727$ & $24.30$ & $70.28$ & $0.8173$ & $46.99$ & $78.11$ & $1.2731$ & $27.11$ & $62.05$ \\
 & T+P &  & $1.2129$ & $27.91$ & $66.67$ & $0.8956$ & $41.37$ & $77.91$ & $0.9679$ & $32.53$ & $77.71$ \\
 & I+P &  & $1.0783$ & $27.71$ & $75.10$ & $0.8072$ & $43.37$ & $81.73$ & $0.9257$ & $34.74$ & $78.51$ \\
\midrule
\multirow{3}{=}{Kimi-K3} & I & \multirow{3}{*}{2.8T} & \cellcolor{aipfArousalBest}$0.8956$ & $35.14$ & \cellcolor{aipfArousalBest}$81.73$ & $0.7470$ & $45.38$ & $84.14$ & $0.9759$ & $33.33$ & $76.71$ \\
 & T+P &  & $1.0281$ & $32.73$ & $73.90$ & $0.8414$ & $42.97$ & $79.12$ & $0.8956$ & $32.73$ & $81.33$ \\
 & I+P &  & $0.9980$ & $33.13$ & $76.91$ & $0.7329$ & $47.99$ & $82.13$ & \cellcolor{aipfValenceBest}$0.8394$ & $36.55$ & \cellcolor{aipfValenceBest}$83.33$ \\
\midrule
\multirow{3}{=}{DeepSeek-4.1-Flash} & I & \multirow{3}{*}{552B} & $0.9799$ & $34.54$ & $76.31$ & $0.7550$ & $47.79$ & $81.12$ & $1.0964$ & $30.92$ & $72.29$ \\
 & T+P &  & $0.9458$ & \cellcolor{aipfArousalBest}$37.35$ & $77.51$ & $0.7169$ & \cellcolor{aipfDominanceBest}$50.40$ & $81.93$ & $0.8835$ & $35.94$ & $79.32$ \\
 & I+P &  & $1.0703$ & $29.72$ & $73.69$ & $0.7289$ & $50.20$ & $81.53$ & $0.8835$ & \cellcolor{aipfValenceBest}$37.95$ & $78.11$ \\
\midrule
GLM-5.3 & T+P & 744B & $2.2229$ & $7.03$ & $24.30$ & $0.7349$ & $48.80$ & $81.93$ & $0.9116$ & $33.94$ & $79.32$ \\
\midrule
\multirow{3}{=}{Qwen3.5-9B (off)} & I & \multirow{3}{*}{9B} & $1.5161$ & $23.69$ & $59.24$ & $1.3434$ & $30.52$ & $59.24$ & $1.5141$ & $21.08$ & $55.42$ \\
 & T+P &  & $1.0763$ & $29.92$ & $73.90$ & $1.2410$ & $31.73$ & $65.46$ & $1.2631$ & $24.90$ & $66.87$ \\
 & I+P &  & $1.2570$ & $27.31$ & $67.87$ & $1.4759$ & $22.89$ & $58.63$ & $1.4378$ & $20.88$ & $58.63$ \\
\midrule
\multirow{3}{=}{Qwen3.5-9B (on)} & I & \multirow{3}{*}{9B} & $1.1888$ & $26.51$ & $70.08$ & $1.0261$ & $42.97$ & $69.48$ & $1.1707$ & $27.11$ & $68.88$ \\
 & T+P &  & $0.9458$ & $35.54$ & $76.10$ & $0.8896$ & $42.77$ & $77.51$ & $0.9779$ & $33.13$ & $75.50$ \\
 & I+P &  & $1.1406$ & $29.72$ & $69.08$ & $0.9779$ & $42.57$ & $73.09$ & $1.0783$ & $31.33$ & $71.08$ \\
\midrule
\multicolumn{12}{@{}l}{\textit{B. Ours}} \\
\addlinespace[1pt]
\multirow{2}{=}{\begingroup\setlength{\fboxsep}{0pt}\colorbox{aipfOursBg}{\parbox[c][2\baselineskip][c]{49pt}{\centering\textbf{\SubJudge}}}\endgroup} & \cellcolor{aipfOursBg}\textbf{T+P} & \cellcolor{aipfOursBg}\textbf{9B} & \cellcolor{aipfArousalHead}$\mathbf{0.7631}$ & \cellcolor{aipfArousalHead}$\mathbf{43.17}$ & \cellcolor{aipfArousalHead}$\mathbf{83.94}$ & \cellcolor{aipfDominanceHead}$\mathbf{0.6627}$ & \cellcolor{aipfDominanceHead}$\mathbf{51.61}$ & \cellcolor{aipfDominanceHead}$\mathbf{84.74}$ & \cellcolor{aipfValenceHead}$\mathbf{0.9297}$ & \cellcolor{aipfValenceHead}$\mathbf{34.34}$ & \cellcolor{aipfValenceHead}$\mathbf{77.11}$ \\
 & \cellcolor{aipfOursBg}\textbf{I+P} & \cellcolor{aipfOursBg}\textbf{9B} & \cellcolor{aipfArousalHead}$\mathbf{0.7711}$ & \cellcolor{aipfArousalHead}$\mathbf{44.18}$ & \cellcolor{aipfArousalHead}$\mathbf{83.33}$ & \cellcolor{aipfDominanceHead}$\mathbf{0.6908}$ & \cellcolor{aipfDominanceHead}$\mathbf{47.19}$ & \cellcolor{aipfDominanceHead}$\mathbf{86.14}$ & \cellcolor{aipfValenceHead}$\mathbf{0.9900}$ & \cellcolor{aipfValenceHead}$\mathbf{31.93}$ & \cellcolor{aipfValenceHead}$\mathbf{75.70}$ \\
\bottomrule
\end{tabular}
\end{lrbox}
\ifdim\wd\aipftablebox>\linewidth
\resizebox{\linewidth}{!}{\usebox{\aipftablebox}}
\else
\usebox{\aipftablebox}
\fi
\endgroup
\end{minipage}
\end{table}
\endgroup
 
\subsection{Pairwise Voting on iNews}
\label{app:inews-pairwise-voting}

Table~\ref{tab:inews-pairwise-votes} summarizes the decisions of five judges
on the same 232{,}009 comparisons per dimension, comprising 89{,}096
same-person comparisons across cases and 142{,}913 same-case comparisons
across people. We count wins, losses, and ties relative to the first
case--person pair in the fixed presentation order and aggregate the five
votes using Eq.~\ref{eq:judge-vote}. The judges differ in their tendency
to declare ties, with Arousal tie rates ranging from 6.59\% to 26.99\%.
The combined vote yields 9{,}489 Arousal ties, 12{,}383 Dominance ties, and
3{,}889 Valence ties, corresponding to 4.09\%, 5.34\%, and 1.68\%
of comparisons, respectively. These rates are lower than those of every
individual judge in the corresponding dimension, providing more strictly
ordered comparisons for scale construction.
\begin{table}[!htbp]
\centering
\caption{Pairwise voting outcomes on iNews. Each judge evaluates the same 232{,}009 pairs in each dimension.
Win and Loss indicate that the first case--person pair in the fixed presentation
order is judged higher or lower on the given dimension than the second, respectively.
Tie denotes a tied judgment. The five-judge vote compares the numbers of Win
and Loss votes, with equal counts yielding a tie, as in Eq.~\ref{eq:judge-vote}.
}
\label{tab:inews-pairwise-votes}
\begingroup
\fontsize{9}{10.5}\selectfont
\setlength{\tabcolsep}{3pt}
\renewcommand{\arraystretch}{1.12}
\begin{lrbox}{\aipftablebox}
\begin{tabular}{@{}l*{3}{>{\columncolor[HTML]{F2F8F5}}r}*{3}{>{\columncolor[HTML]{F8F4FB}}r}*{3}{>{\columncolor[HTML]{F3F7FC}}r}@{}}
\toprule
\multirow{2}{*}{Judge}
& \multicolumn{3}{>{\columncolor[HTML]{E0EEE7}}c}{\textbf{Arousal}}
& \multicolumn{3}{>{\columncolor[HTML]{ECE3F4}}c}{\textbf{Dominance}}
& \multicolumn{3}{>{\columncolor[HTML]{E1EBF7}}c}{\textbf{Valence}} \\
\cmidrule(lr){2-4}\cmidrule(lr){5-7}\cmidrule(l){8-10}
& Win & Loss & Tie & Win & Loss & Tie & Win & Loss & Tie \\
\midrule
GPT-5.4 & 109,735 & 103,559 & 18,715 & 94,956 & 97,294 & 39,759 & 112,993 & 111,376 & 7,640 \\
GPT-5.5 & 103,919 & 108,273 & 19,817 & 94,709 & 102,772 & 34,528 & 109,209 & 111,625 & 11,175 \\
GPT-5.6-sol & 109,519 & 107,208 & 15,282 & 99,057 & 103,129 & 29,823 & 109,504 & 112,640 & 9,865 \\
Gemini-2.5-Pro & 81,426 & 87,964 & 62,619 & 96,245 & 116,711 & 19,053 & 112,302 & 114,008 & 5,699 \\
Gemini-3.5-Flash & 102,857 & 100,820 & 28,332 & 99,481 & 97,751 & 34,777 & 113,535 & 105,340 & 13,134 \\
\midrule
\textbf{Five-judge vote} & 111,386 & 111,134 & 9,489 & 106,175 & 113,451 & 12,383 & 114,018 & 114,102 & 3,889 \\
\bottomrule
\end{tabular}
\end{lrbox}
\ifdim\wd\aipftablebox>\linewidth
  \resizebox{\linewidth}{!}{\usebox{\aipftablebox}}
\else
  \usebox{\aipftablebox}
\fi
\endgroup
\end{table}

\clearpage
\subsection{Ablation on Elo}
\label{app:elo-ablation}

We assess the sensitivity of Elo scale construction to replay order,
update step $\eta$, scale parameter $\tau$, and initial rating $r_0$
using the fixed five-judge comparisons on \textsc{PluriHarms}.
In Table~\ref{tab:pluriharms-elo-ablation}, pooled agreement varies by less than one
percentage point, from 78.020\% to 78.964\%. At the default parameters,
ten global shuffles give 78.756\% agreement with a standard deviation of
0.026 percentage points, indicating little sensitivity to random replay order.
A common initial rating of 0, 1500, or 3000 produces identical results
in all categories, since a uniform shift preserves Elo differences.

\begin{table}[!htbp]
\centering
\caption{Elo parameter and order ablation on \textsc{PluriHarms}.
Global agreement covers all 45{,}583{,}499 pairs with different human ratings
among the 9{,}789 alignment objects. Concordant, discordant, and tied
Elo orderings receive 1, 0, and 0.5. All settings use the same five-judge
votes, a budget of 20 opponents per axis, and one replay ($T=1$).
Orders counts sequences. Global shuffle reports mean $\pm$ standard
deviation over seeds 0--9, in percentage points. Within-round random
uses seed 42. Bold marks the default configuration, and darker cells
mark the highest mean among global-shuffle settings.}
\label{tab:pluriharms-elo-ablation}
\begingroup
\fontsize{9}{10.5}\selectfont
\setlength{\tabcolsep}{3pt}
\renewcommand{\arraystretch}{1.15}
\begin{tabular*}{\linewidth}{@{\extracolsep{\fill}}lccccc>{\columncolor[HTML]{F2F8F5}}c@{}}
\toprule
Replay order & $r_0$ & $\eta$ & $\tau$ & $T$ & Orders
& \cellcolor[HTML]{E0EEE7}\shortstack{Global agreement\\(\%) $\uparrow$} \\
\midrule
\textbf{Within-round random} & \textbf{1500} & \textbf{32} & \textbf{400} & \textbf{1} & \textbf{1} & \cellcolor[HTML]{E0EEE7}$\boldsymbol{78.819}$ \\
Global shuffle & 1500 & 32 & 400 & 1 & 10 & $78.756 \pm 0.026$ \\
Reverse & 1500 & 32 & 400 & 1 & 1 & $78.708$ \\
Same-person first & 1500 & 32 & 400 & 1 & 1 & $78.569$ \\
Same-prompt first & 1500 & 32 & 400 & 1 & 1 & $78.593$ \\
\midrule
Global shuffle & 1500 & 8 & 400 & 1 & 10 & $78.020 \pm 0.009$ \\
Global shuffle & 1500 & 16 & 400 & 1 & 10 & $78.391 \pm 0.015$ \\
Global shuffle & 1500 & 64 & 400 & 1 & 10 & \cellcolor[HTML]{C8E2D3}$78.964 \pm 0.039$ \\
Global shuffle & 1500 & 128 & 400 & 1 & 10 & $78.925 \pm 0.060$ \\
\midrule
Global shuffle & 1500 & 32 & 200 & 1 & 10 & \cellcolor[HTML]{C8E2D3}$78.964 \pm 0.039$ \\
Global shuffle & 1500 & 32 & 800 & 1 & 10 & $78.391 \pm 0.015$ \\
\midrule
Within-round random & 0 & 32 & 400 & 1 & 1 & $78.819$ \\
Within-round random & 3000 & 32 & 400 & 1 & 1 & $78.819$ \\
\bottomrule
\end{tabular*}
\endgroup
\end{table}

Table~\ref{tab:pluriharms-elo-pair-types} separately decomposes the default
configuration by pair type. For the same person across different prompts
(intensity), agreement is 75.659\%. For the same prompt across different
people (individual variation), it is 72.959\%. Different people with
different prompts (cross pairs) yield 78.906\%. Each group includes all
eligible pairs of observed objects, beyond those directly compared during
scale construction. Cross pairs have no direct comparison edges, so their
relative order is inferred indirectly through the connected Elo graph.
They account for 98.12\% of the non-tied pairs and therefore dominate
the pooled global agreement of 78.819\%.
 
\begin{table}[!htbp]
\centering
\caption{Elo agreement by pair type under the default configuration
($r_0=1500$, $\eta=32$, $\tau=400$, $T=1$, within-round random with seed 42).
Each row evaluates all unordered object pairs of its type using the same
final Elo scores, including pairs that never directly competed.
GT ties are excluded from Valid pairs. For a person with $n_p$ observed
prompts, the same-person count before excluding ties is $\binom{n_p}{2}$,
summed over people. Same-prompt counts are computed analogously over prompts.
Agreement gives 1 for concordant rankings, 0 for discordant rankings,
and 0.5 for Elo ties.}
\label{tab:pluriharms-elo-pair-types}
\begingroup
\fontsize{9}{10.5}\selectfont
\setlength{\tabcolsep}{3pt}
\renewcommand{\arraystretch}{1.18}
\begin{tabular*}{\linewidth}{@{\extracolsep{\fill}}lrrrc@{}}
\toprule
Pair type & All pairs & GT ties & Valid pairs
& \shortstack{Agreement\\(\%) $\uparrow$} \\
\midrule
Same person, different prompts & 475{,}430 & 47{,}421 & 428{,}009 & \cellcolor[HTML]{F3F7FC}$75.659$ \\
Same prompt, different people & 474{,}510 & 45{,}433 & 429{,}077 & \cellcolor[HTML]{F8F4FB}$72.959$ \\
Different people and prompts & 46{,}957{,}426 & 2{,}231{,}013 & 44{,}726{,}413 & \cellcolor[HTML]{F2F8F5}$78.906$ \\
\midrule
\textbf{All pairs} & \textbf{47{,}907{,}366} & \textbf{2{,}323{,}867} & \textbf{45{,}583{,}499} & \cellcolor[HTML]{E0EEE7}$\boldsymbol{78.819}$ \\
\bottomrule
\end{tabular*}
\endgroup
\end{table}

\subsection{Pairwise Comparison vs.\ Direct Scoring}
\label{app:pairwise-vs-direct}

Instead of comparing two records, a judge can score each record
directly.
We collect such direct scores on the 4{,}926 iNews target records used
for teacher comparisons, with the same screenshot and profile inputs.
Table~\ref{tab:inews-direct-vs-pairwise} compares the two protocols for
GPT-5.4 and Gemini-3.5-Flash, two judges from our panel, on the 4{,}897
records with valid direct scores from every judge. 
The five-judge
scale achieves the highest agreement in every column, exceeding direct
scoring by 2.0 to 5.6 points over all pairs and by 1.7 to 4.9 points on
pairs whose human ratings differ by one point. These results show that direct scoring falls short of our pairwise
comparison, even between adjacent ratings.

\begin{table}[!htbp]
\centering
\caption{Pairwise ranking agreement (\%) with human ratings on iNews over the 4{,}897 target records with valid direct scores. All covers every pair of records with different human ratings, and $\Delta{=}1$ keeps pairs one rating point apart. Concordant, discordant, and tied predictions receive 1, 0, and 0.5. Direct scoring rates each record from 1 to 7.}
\label{tab:inews-direct-vs-pairwise}
\begingroup
\definecolor{aipfArousalBg}{HTML}{F2F8F5}
\definecolor{aipfArousalHead}{HTML}{E0EEE7}
\definecolor{aipfArousalBest}{HTML}{C8E2D3}
\definecolor{aipfArousalInk}{HTML}{356D57}
\definecolor{aipfDominanceBg}{HTML}{F8F4FB}
\definecolor{aipfDominanceHead}{HTML}{ECE3F4}
\definecolor{aipfDominanceBest}{HTML}{DCCCEB}
\definecolor{aipfDominanceInk}{HTML}{725391}
\definecolor{aipfValenceBg}{HTML}{F3F7FC}
\definecolor{aipfValenceHead}{HTML}{E1EBF7}
\definecolor{aipfValenceBest}{HTML}{CADCF1}
\definecolor{aipfValenceInk}{HTML}{355D8A}
\definecolor{aipfOursBg}{HTML}{EDF0F5}
\fontsize{8}{9.2}\selectfont
\setlength{\tabcolsep}{2pt}
\renewcommand{\arraystretch}{1.08}
\begin{lrbox}{\aipftablebox}
\begin{tabular}{@{}>{\raggedright\arraybackslash}p{58pt}>{\raggedright\arraybackslash}p{64pt}*{2}{>{\columncolor{aipfArousalBg}\centering\arraybackslash}p{38pt}}*{2}{>{\columncolor{aipfDominanceBg}\centering\arraybackslash}p{38pt}}*{2}{>{\columncolor{aipfValenceBg}\centering\arraybackslash}p{38pt}}@{}}
\toprule
\multirow{2}{*}{Method} & \multirow{2}{*}{Model}
& \multicolumn{2}{>{\columncolor{aipfArousalHead}}c}{\textcolor{aipfArousalInk}{\textbf{Arousal}}}
& \multicolumn{2}{>{\columncolor{aipfDominanceHead}}c}{\textcolor{aipfDominanceInk}{\textbf{Dominance}}}
& \multicolumn{2}{>{\columncolor{aipfValenceHead}}c}{\textcolor{aipfValenceInk}{\textbf{Valence}}} \\
\cmidrule(lr){3-4}\cmidrule(lr){5-6}\cmidrule(l){7-8}
& & All $\uparrow$ & $\Delta{=}1$ $\uparrow$ & All $\uparrow$ & $\Delta{=}1$ $\uparrow$ & All $\uparrow$ & $\Delta{=}1$ $\uparrow$ \\
\midrule
\multirow{2}{*}{Direct scoring} & GPT-5.4 & $68.32$ & $61.20$ & $70.61$ & $63.45$ & $75.51$ & $65.62$ \\
 & Gemini-3.5-Flash & $68.49$ & $60.60$ & $69.61$ & $62.83$ & $76.49$ & $66.35$ \\
\midrule
\multirow{2}{*}{Pairwise} & GPT-5.4 & $70.09$ & $62.70$ & $74.60$ & $67.21$ & $77.09$ & $66.86$ \\
 & Gemini-3.5-Flash & $70.58$ & $63.32$ & $73.82$ & $66.55$ & $76.53$ & $66.30$ \\
\midrule
\cellcolor{aipfOursBg}\textbf{Pairwise } & \cellcolor{aipfOursBg}\textbf{Five judges}
& \cellcolor{aipfArousalBest}$\mathbf{70.59}$ & \cellcolor{aipfArousalBest}$\mathbf{63.40}$
& \cellcolor{aipfDominanceBest}$\mathbf{75.20}$ & \cellcolor{aipfDominanceBest}$\mathbf{67.70}$
& \cellcolor{aipfValenceBest}$\mathbf{78.49}$ & \cellcolor{aipfValenceBest}$\mathbf{68.05}$ \\
\bottomrule
\end{tabular}
\end{lrbox}
\ifdim\wd\aipftablebox>\linewidth
\resizebox{\linewidth}{!}{\usebox{\aipftablebox}}
\else
\usebox{\aipftablebox}
\fi
\endgroup
\end{table}

\section{PluriHarms Efficiency Details}
\label{sec:plu-efficiency-details}

\textbf{Inputs and record selection.}
\SubJudge timings use the reference configuration, which is trained on the
original five-judge panel with BPO only and selected at checkpoint 876, with
its LoRA merged into the backbone. The main configuration shares the same
architecture and readout.
The workload uses the original \textsc{PluriHarms} API user message for each held-out prompt. Target human ratings do not
enter inference, timing, implementation selection, or record selection.
We retain all 50 distinct test questions. For each question and
configuration, one available sequential measurement is selected uniformly
with seed 20260920, before reading target ratings for offline evaluation.
\SubJudge and thinking off have three available measurements per question.
Thinking on has two measurements for 42 questions and one for eight.
Predicted scores agree across available repeats for every question.
Statistics describe these 50 selected requests, without claiming a
between-run standard deviation from three complete repetitions.

\textbf{Generation and timing.}
The backbone uses greedy decoding. Thinking on has a 4,096-token reasoning
budget. If the reasoning block remains open, three delimiter tokens
representing a newline, \texttt{</think>} and a trailing blank line are
injected within the same generation call, retaining the KV cache. The
answer has no separate length cap and must terminate with EOS. Only text
after the closing token is parsed. Among the selected requests, 49 require
forced closure and one closes naturally. All return a score, with five
answer tokens on average and no exhaustion of the model context capacity.
The response-position count includes both generated and injected tokens.

The \SubJudge and thinking-off records come from the preceding completed
configuration measurements on the same inputs, hardware class, model
assets and software runtime. Thinking off used a 32-token cap, which none
of its selected requests reaches: the maximum is six tokens. \SubJudge uses
the same merged checkpoint and digit readout.
One complete batch-8 repetition is selected uniformly for each of these 
two configurations, retaining its seven original batches and all 50
questions. Thinking on is measured separately on the same 50 questions
in one batch-8 pass, comprising six batches of eight and a final batch
of two. It yields 50 valid scores, with 50 forced reasoning closures
and 0 cases of exhausted model context. The total measured batch
time is 1190.255 seconds. This supplemental measurement
uses the same generation and timing code as the sequential experiment.

\textbf{256-token thinking budget.}
Two complete passes use the same 50 questions, model weights, tokenized
inputs, BF16/SDPA runtime, and greedy decoding, with the thinking budget
reduced to 256 tokens. The batch-1 pass takes 440.453 seconds in total.
Response positions average 264.00 and reach 265. Mean latency is 8.809
seconds, with P95 8.896 seconds and observed maximum 8.951 seconds. All
50 requests require forced closure, end at natural EOS, and avoid context
exhaustion. The separate seven-batch batch-8 pass takes 81.108 seconds, 
yielding 0.616 valid requests/s and 24.74 GiB peak allocated memory. The
throughput ratio relative to the 4,096-token pass is $14.67\times$,
computed before rounding. Each pass is a single complete run without an 
across-run standard deviation. The same timing exclusions apply.
 
\textbf{Measurement protocol.}
Response length and latency are computed from 50 sequential requests at
batch size 1. For the backbone, response length includes reasoning,
answer, EOS, and injected closing tokens. For \SubJudge, it denotes one
scored response position without token generation. Throughput and memory
are measured separately with a single batch of 64 requests, formed by
repeating the first 14 inputs after the original 50. Throughput is the
number of valid scores divided by total measured time, and memory is
peak allocated GPU memory including model weights. Timing includes
preprocessing, device transfer, inference, forced closure, and score
extraction, while excluding model loading and warmup. All sequential and
batched requests yield valid scores.

Each batch-64 configuration is measured once. Every thinking-on request
uses forced closure followed by natural EOS, without an answer-length
cap or context exhaustion. Input-file preloading and logging are also excluded from timing.
There is no cross-request prefix cache.
The reported latency maxima are observed values. The digit-only readout
was selected on alignment inputs before test timing and agrees with the
original merged readout within the prescribed numerical tolerance.

\begin{table}[!htbp]
\centering
\caption{Sequential timing breakdown for the selected 50 requests. First-position time includes preprocessing and device transfer. Later decoding includes every subsequent generation step. Completion counts EOS-terminated valid scores for the backbone and valid direct scores for \SubJudge.}
\small
\begin{tabular}{@{}lrrrr@{}}
\toprule
Configuration & First position (s) & Later decode (s) & Completion (\%) & Forced close (\%) \\
\midrule
Qwen3.5-9B, off & 0.508 & 0.140 & 100.0 & 0.0 \\
Qwen3.5-9B, on & 0.485 & 130.688 & 100.0 & 98.0 \\
\SubJudge, merged LoRA & 0.503 & 0.000 & 100.0 & 0.0 \\
\bottomrule
\end{tabular}
\end{table}

\end{document}